\documentclass{article}
\usepackage{spconf,amsmath,graphicx}

\usepackage[hyphens]{url}
\usepackage[colorlinks=true,allcolors=black]{hyperref}
\usepackage{textcomp} 
\usepackage{gensymb} 
\usepackage{arydshln}
\usepackage{booktabs}
\usepackage{multirow}
\usepackage[utf8]{inputenc}
\usepackage[acronym]{glossaries}
\usepackage[caption=false,farskip=0pt]{subfig}
\usepackage[table,dvipsnames]{xcolor}
\usepackage{xspace} 
\usepackage[sort&compress,numbers]{natbib}
\usepackage{balance}
\usepackage{diagbox}
\usepackage{enumitem}
\newcommand{\minus}{\scalebox{0.75}[1.0]{$-$}}

\newcommand\colored[1]{#1} 
\newcommand\minor[1]{#1} %

\newlength{\adj}
\newacronymstyle{long-short-br}
{%
  \GlsUseAcrEntryDispStyle{long-short}%
}%
{%
  \GlsUseAcrStyleDefs{long-short}%
}
\setacronymstyle{long-short-br}

\title{\Large An Efficient and Layout-Independent Automatic License Plate\\Recognition System Based on the YOLO detector}

\makeatletter
\def\@name{ \emph{Rayson Laroca\textsuperscript{1}, Luiz~A.~Zanlorensi\textsuperscript{1}, Gabriel~R.~Gon\c{c}alves\textsuperscript{2},}  \\[0.1ex] \emph{Eduardo~Todt\textsuperscript{1}, William~Robson~Schwartz\textsuperscript{2}, David~Menotti\textsuperscript{1}}\thanks{
\scriptsize{This paper is a postprint of a paper accepted by \emph{IET Intelligent Transport Systems} and is subject to Institution of Engineering and Technology Copyright. The copy of record is available at the \emph{Wiley Online~Library} (DOI: \href{http://doi.org/10.1049/itr2.12030}{\textcolor{blue}{10.1049/itr2.12030}}).}} \\}
\makeatother

\address{\textsuperscript{1}Department of Informatics, Federal University of Paran\'a, Curitiba, Brazil\\\textsuperscript{2}Department of Computer Science, Federal University of Minas Gerais, Belo Horizonte, Brazil\\[0.5ex] \textsuperscript{1}\textit{\{rblsantos, lazjunior, todt, menotti\}@inf.ufpr.br} \quad \textsuperscript{2}\textit{\{gabrielrg, william\}@dcc.ufmg.br}}

\begin{document}
\ninept
\sloppy
\maketitle
\newacronym{ac}{AC}{access control}
\newacronym{alpr}{ALPR}{Automatic License Plate Recognition}
\newacronym{anpr}{ANPR}{Automatic Number Plate Recognition}
\newacronym{aolp}{AOLP}{Application-Oriented License Plate}
\newacronym{ap}{AP}{Average Precision}
\newacronym{api}{API}{Application Programming Interface}
\newacronym{flop}{FLOP}{floating-point operations}
\newacronym{bflop}{BFLOP}{billion floating-point operations}
\newacronym{brnn}{BRNN}{Bidirectional Recurrent Neural Network}
\newacronym{cca}{CCA}{Connected Component Analysis}
\newacronym{ccpd}{CCPD}{Chinese City Parking Dataset}
\newacronym{cnn}{CNN}{Convolutional Neural Network}
\newacronym{crnn}{CRNN}{Convolutional Recurrent Neural Network}
\newacronym{coco}{COCO}{Common Objects in Context}
\newacronym{ctc}{CTC}{Connectionist Temporal Classification}
\newacronym{elm}{ELM}{Extreme Learning Machine}
\newacronym{fn}{FN}{false negative}
\newacronym{fp}{FP}{false positive}
\newacronym{fpn}{FPN}{Feature Pyramid Network}
\newacronym{fps}{FPS}{frames per second}
\newacronym{gan}{GAN}{Generative Adversarial Network}
\newacronym{gpu}{GPU}{Graphics Processing Unit}
\newacronym{hmm}{HMM}{Hidden Markov Model}
\newacronym{ilsvrc}{ILSVRC}{ImageNet Large Scale Visual Recognition Challenge}
\newacronym{its}{ITS}{Intelligent Transportation Systems}
\newacronym{iou}{IoU}{Intersection over Union}
\newacronym{le}{LE}{law enforcement}
\newacronym{lp}{LP}{license plate}
\newacronym{lpr}{LPR}{License Plate Recognition}
\newacronym{lstm}{LSTM}{Long Short-Term Memory}
\newacronym{map}{mAP}{mean Average Precision}
\newacronym{mercosur}{Mercosur}{\textit{Mercado Común del Sur} - Southern Common Market}
\newacronym{mlp}{MLP}{Multilayer Perceptron}
\newacronym{mnist}{MNIST}{Modified National Institute of Standards and Technology}
\newacronym{nms}{NMS}{Non-Maximum Suppression}
\newacronym{ocr}{OCR}{Optical Character Recognition}
\newacronym{png}{PNG}{Portable Network Graphics}
\newacronym{prelu}{PReLU}{Parametric ReLU}
\newacronym{rcnn}{RCNN}{Region-based CNN}
\newacronym{relu}{ReLU}{Rectified Linear Unit}
\newacronym{resnet}{ResNet}{Residual Network}
\newacronym{rp}{RP}{road patrol}
\newacronym{rpn}{RPN}{Region Proposal Network}
\newacronym{roi}{ROI}{Region of Interest}
\newacronym{snow}{SNoW}{Sparse Network of Winnows}
\newacronym{smqt}{SMQT}{Successive Mean Quantization Transform}
\newacronym{ssd}{SSD}{Single Shot MultiBox Detector}
\newacronym{svm}{SVM}{Support Vector Machine}
\newacronym{tn}{TN}{true Negative}
\newacronym{tp}{TP}{true Positive}
\newacronym{voc}{VOC}{Visual Object Classes}
\newacronym{wpodnet}{WPOD-NET}{Warped Planar Object Detection Network}
\newacronym{yolo}{YOLO}{You Only Look Once}

\newcommand{\dataset}{UFPR-ALPR\xspace}

\newcommand{\aolp}{AOLP\xspace}
\newcommand{\aolpe}{AOLPE\xspace}
\newcommand{\caltech}{Caltech Cars\xspace}
\newcommand{\ccpd}{CCPD\xspace}
\newcommand{\cdhard}{CD-HARD\xspace}
\newcommand{\chinese}{ChineseLP\xspace}
\newcommand{\englishlpd}{EnglishLP\xspace}
\newcommand{\ircp}{IRCP\xspace}
\newcommand{\medialab}{Medialab LPR\xspace}
\newcommand{\ntustmlp}{NTUST-MLP\xspace}
\newcommand{\openalpreu}{OpenALPR-EU\xspace}
\newcommand{\pku}{PKU\xspace}
\newcommand{\ssig}{SSIG-SegPlate\xspace}
\newcommand{\ssigalpr}{SSIG-ALPR\xspace}
\newcommand{\stills}{UCSD-Stills\xspace}
\newcommand{\ucsd}{UCSD\xspace}
\newcommand{\boundingboxes}{38{,}351\xspace}
\newcommand{\images}{6{,}239\xspace}
\newcommand{\supplementary}{https://web.inf.ufpr.br/vri/publications/layout-independent-alpr/}

\newcommand{\accavg}{96.9\%\xspace}
\newcommand{\accavgmath}{$96.9$\%\xspace}
\newcommand{\accufpr}{$90$\%\xspace}

\newcommand{\improvelayoutclassification}{$2.1$\%\xspace}

\newcommand{\outsighthound}{$9.1$\%\xspace}
\newcommand{\outopenalpr}{$6.2$\%\xspace}
%

\begin{abstract}
\textit{This paper presents an efficient and layout‐independent \gls*{alpr} system based on the state‐of‐the‐art YOLO object detector that contains a unified approach for \gls*{lp} detection and layout classification to improve the recognition results using post‐processing rules.
The system is conceived by evaluating and optimizing different models, aiming at achieving the best speed/accuracy trade‐off at each stage.
The networks are trained using images from several datasets, with the addition of various data augmentation techniques, so that they are robust under different conditions. The proposed system achieved an average end‐to‐end recognition rate of~\accavg across eight public datasets (from five different regions) used in the experiments, outperforming both previous works and commercial systems in the \chinese, \openalpreu, \ssig and \dataset datasets.
In the other datasets, the proposed approach achieved competitive results to those attained by the baselines.
Our system also achieved impressive \gls*{fps} rates on a high-end~\acrshort*{gpu}, being able to perform in real time even when there are four vehicles in the scene. An additional contribution is that we manually labeled \boundingboxes bounding boxes on \images images from public datasets and made the annotations publicly available to the research community.}
\end{abstract}
%
%
\section{Introduction}
\label{sec:introduction}

\glsresetall

\gls*{alpr} became an important topic of research since the appearance of the first works in the early $1990$s~\citep{lotufo1990automatic,kanayama1991development}.
A variety of \gls*{alpr} systems and commercial products have been produced over the years due to many practical applications such as automatic toll collection, border control, traffic law enforcement and road traffic monitoring~\citep{anagnostopoulos2008license,du2013automatic}.

\gls*{alpr} systems typically include three phases, namely: \gls*{lp} detection, character segmentation and character recognition, which refer to (i)~locating the \gls*{lp} region in the acquired image, (ii)~segmenting each character within the detected \gls*{lp} and (iii)~classifying each segmented character.
The earlier stages require higher accuracy since a failure would probably lead to another failure in the subsequent stages.

Many authors have proposed approaches with a vehicle detection stage prior to \gls*{lp} detection, aiming to eliminate \glspl*{fp} and reduce processing time~\citep{goncalves2016license,silva2017realtime,silva2018license}.
Regarding character segmentation, it has become common the use of segmentation-free approaches for \gls*{lp} recognition~\citep{goncalves2018realtime,bulan2017segmentation,spanhel2017holistic,goncalves2019multitask}, as the character segmentation by itself is a challenging task that is prone to be influenced by uneven lighting conditions, shadows, and noise~\citep{li2018reading}.

Despite the major advances (in terms of both accuracy and efficiency) that have been achieved in computer vision using deep learning~\citep{lecun2015deep}, several solutions are still not robust enough to be executed in real-world scenarios.
Such solutions commonly depend on certain constraints such as specific cameras or viewing angles, simple backgrounds, good lighting conditions, search in a fixed region, and certain types of vehicles.
Additionally, many authors still propose computationally expensive approaches that are not able to process frames in real time, even when the experiments are performed on a high-end \acrshort*{gpu}~\citep{dong2017cnnbased,li2018reading,li2018toward}
\minor{(if a system does not perform in real time using a high-end GPU, it is very unlikely to run fast enough on the mid-end GPUs that are often employed in real-world applications).}
In the literature, generally a system is considered ``real-time'' if it can process at least $30$~\gls*{fps} since commercial cameras usually record videos at that frame rate~\citep{redmon2016yolo,goncalves2018realtime,laroca2018robust}.

\gls*{alpr} systems must also be capable of recognizing multiple \gls*{lp} layouts since there might be various \gls*{lp} layouts in the same country or region.
However, as stated in~\cite{kessentini2019twostage}, most of the existing solutions work only for a specific \gls*{lp} layout.
Even though most authors claim that their approaches could be extended with small modifications to detect/segment/recognize \glspl*{lp} of different layouts~\citep{tian2015twostage,gou2016vehicle,dong2017cnnbased,yuan2017robust}, this may not be an easy task.
For instance, a character segmentation approach designed for \glspl*{lp} with simple backgrounds is likely to fail on \glspl*{lp} with complex backgrounds and logos that touch and overlap some characters (e.g., Florida \glspl*{lp})~\citep{bulan2017segmentation}.

\minor{A robust and efficient \gls*{alpr} system can play a key role in various applications in the context of intelligent transportation systems.
For example, vehicle re-identification, which refers to identifying a target vehicle in different cameras with non-overlapping views~\cite{khan2019survey}, is known to be a very challenging problem since different vehicles with the same model and color are highly similar to each other.
Although in these situations the \gls*{lp} information must be considered for precise vehicle search~\cite{liu2018provid,oliveira2019vehicle}, it is generally not explored due to the limitations of existing~\gls*{alpr} systems in unconstrained~scenarios~\cite{he2019part,hou2019deep}.}

\minor{Considering the above discussion,} we propose an end-to-end, efficient and layout-independent \gls*{alpr} system exploring YOLO-based models at all stages.
YOLO~\citep{redmon2016yolo,redmon2017yolo9000,redmon2018yolov3} is a real-time object detector that achieved impressive results in terms of speed/accuracy trade-off in the Pascal~\acrshort*{voc}~\citep{everingham2010pascalvoc} and Microsoft~\acrshort*{coco}~\citep{lin2014microsoft} detection tasks.
We locate the vehicles in the input image and then their \glspl*{lp} within the vehicle bounding box.
Considering that the bottleneck of \gls*{alpr} systems is the \gls*{lp} recognition stage (see Section~\ref{sec:related_work:final_remarks}), in this paper we propose a unified approach for \gls*{lp} detection and \textit{layout classification} to improve the recognition results using post-processing rules (this is the first time a layout classification stage is proposed to improve \gls*{lp} recognition, to the best of our knowledge).
Afterward, all \gls*{lp} characters are recognized simultaneously, i.e., the entire \gls*{lp} patch is fed into the network, avoiding the challenging character segmentation~task.

We eliminate various constraints commonly found in \gls*{alpr} systems by training a single network for each task using images from several datasets, which were collected under different conditions and reproduce distinct real-world applications.
Moreover, we perform several data augmentation tricks and modified the chosen networks \minor{(e.g., we explored various models with changes in the input size, as well as in the number of layers, filters, output classes, and anchors)} aiming to achieve the best speed/accuracy trade-off at each~stage.

Our experimental evaluation demonstrates the effectiveness of the proposed approach, which outperforms previous works and two commercial systems in the \chinese~\citep{zhou2012principal}, \openalpreu~\citep{openalpreu}, \ssig~\citep{goncalves2016benchmark} and \dataset~\citep{laroca2018robust} datasets, and achieves competitive results to those attained by the baselines in other four public~datasets.
Our system also achieved an impressive trade-off between accuracy and speed. Specifically, on a high-end \acrshort*{gpu} (i.e., an NVIDIA Titan~XP), the proposed system is able to process images in real time even when there are $4$ vehicles in the~scene.

\minor{In summary, the main contributions of this work are: 
\begin{itemize}[labelsep=1mm]
    \item A new \emph{\textbf{end-to-end}}, \emph{\textbf{efficient}} and \emph{\textbf{layout-independent}} \gls*{alpr} system that explores YOLO-based \glspl*{cnn} at all stages\footnote{\scriptsize The entire \gls*{alpr} system, i.e., the architectures and weights, along with all annotations made by us are \textbf{publicly available} at \url{\supplementary}.}.
    \begin{itemize}[labelsep=1mm,leftmargin=2mm]
        \item LP~layout classification (along with heuristic rules) greatly improves the recognition results and also enables our system to be easily adjusted for additional/different LP~layouts.
        \item As the proposed \gls*{alpr} system processes more than $70$~\gls*{fps} on a high-end GPU, we believe it can be deployed even in mid-end setups/GPUs for several real-world~applications.
    \end{itemize}
    \item A comparative and detailed evaluation of our approach, previous works in the literature, and two commercial systems in \textbf{eight} publicly available~datasets that have been frequently used to train and/or evaluate algorithms in the \gls*{alpr}~context.
    \begin{itemize}[labelsep=1mm,leftmargin=2mm]
        \item We are not aware of any work in the literature where so many public datasets were used in the~experiments.
    \end{itemize}
    \item Annotations regarding the position of the vehicles, \glspl*{lp} and characters, as well as their classes, in each image of the public datasets used in this work that have no annotations or contain labels only for part of the \gls*{alpr} pipeline. Precisely, we manually labeled $\boundingboxes$ bounding boxes on $\images$ images.
    \begin{itemize}[labelsep=1mm,leftmargin=2mm]
        \item These annotations will considerably assist the development and evaluation of new \gls*{alpr} approaches, as well as the fair comparison among published~works.
    \end{itemize}
\end{itemize}
}

A preliminary version of the system described in this paper was published at the 2018 International Joint Conference on Neural Networks~(IJCNN)~\citep{laroca2018robust}.
The approach described here differs from that version in several aspects.
For instance, in the current version, the~\gls*{lp}~layout is classified prior to \gls*{lp}~recognition (together with \gls*{lp} detection), the recognition of all characters is performed simultaneously (instead of first segmenting and then recognizing each of them) and modifications were made to all networks (e.g., in the input size, number of layers, filters, and anchors, among others) to make them faster and more robust.
In this way, we overcome the limitations of the system presented in~\citep{laroca2018robust} and were able to considerably improve both the execution time (from $28$ms to $14$ms) and the recognition results (e.g., from $64.89$\% to \accufpr in the \dataset dataset).
This version was also evaluated on a broader and deeper~manner. 

The remainder of this paper is organized as follows. We review related works in Section~\ref{sec:related_work}. The proposed system is presented in Section~\ref{sec:proposed}.
In Section~\ref{sec:experiments}, the experimental setup is thoroughly described.
We report and discuss the results in Section~\ref{sec:results}. Finally, conclusions and future works are given in Section~\ref{sec:conclusions}.
\section{Related Work}
\label{sec:related_work}

In this section, we review recent works that use deep learning approaches in the context of \gls*{alpr}. 
For relevant studies using conventional image processing techniques, please refer to~\citep{anagnostopoulos2008license,du2013automatic}.
We first discuss works related to the \gls*{lp} detection and recognition stages, and then conclude with final remarks.

\subsection{License Plate Detection}
\label{sec:related_work:lp_detection}

Many authors have addressed the \gls*{lp} detection stage using object detection \glspl*{cnn}.
Silva \& Jung~\cite{silva2017realtime} noticed that the Fast-YOLO model~\cite{redmon2016yolo} achieved a low recall rate when detecting \glspl*{lp} without prior vehicle detection.
Therefore, they used the Fast-YOLO model arranged in a cascaded manner to first detect the frontal view of the cars and then their \glspl*{lp} in the detected patches, attaining high precision and recall rates on a dataset with Brazilian~\glspl*{lp}.

Hsu et al.~\cite{hsu2017robust} customized the YOLO and YOLOv2 models exclusively for \gls*{lp} detection.
Despite the fact that the modified versions of YOLO performed better and were able to process $54$ \gls*{fps} on a high-end \acrshort*{gpu}, we believe that \gls*{lp} detection approaches should be even faster (i.e., $150+$~\gls*{fps}) since the \gls*{lp} characters still need to be recognized.
Kurpiel et al.~\cite{kurpiel2017convolutional} partitioned the input image in sub-regions, forming an overlapping grid. A score for each region was produced using a \gls*{cnn} and the \glspl*{lp} were detected by analyzing the outputs of neighboring sub-regions.
On a GT-740M \acrshort*{gpu}, it took $230$~ms to detect Brazilian \glspl*{lp} in images with multiple vehicles, achieving a recall rate of $83$\% on a public dataset introduced by~them.

Li et al.~\cite{li2018reading} trained a \gls*{cnn} based on characters cropped from general text to perform a character-based \gls*{lp} detection. 
The network was employed in a sliding-window fashion across the entire image to generate a text salience map.
Text-like regions were extracted based on the clustering nature of the characters. 
\gls*{cca} is subsequently applied to produce the initial candidate boxes.
Then, another \gls*{lp}/non-\gls*{lp} \gls*{cnn} was trained to remove \glspl*{fp}.
Although the precision and recall rates obtained were higher than those achieved in previous works, such a sequence of methods is too expensive for real-time applications, taking more than $2$ seconds to process a single image when running on a Tesla K40c~\acrshort*{gpu}.

Xie et al.~\cite{xie2018new} proposed a YOLO-based model to predict the \gls*{lp} rotation angle in addition to its coordinates and confidence value.
Prior to that, another \gls*{cnn} was applied to determine the attention region in the input image, assuming that some distance will inevitably exist between any two \glspl*{lp}. By cascading both models, their approach outperformed all baselines in three public datasets, while still running in real time. Despite the impressive results, it is important to highlight two limitations in their work: (i) the authors simplified the problem by forcing their \gls*{alpr} system to output only one bounding box per image; (ii) motorcycle \glspl*{lp} might be lost when determining the attention region since, in some scenarios (e.g., traffic lights), they might be very close.
Kessentini et al.~\cite{kessentini2019twostage} detected the \gls*{lp} directly in the input image using YOLOv2 without any change or refinement. However, they also considered only one \gls*{lp} per image (mainly to eliminate false positives in the background), which makes their approach unsuitable for many real-world applications that contain multiple vehicles in the scene~\cite{hsu2017robust,kurpiel2017convolutional,goncalves2018realtime}.

\subsection{License Plate Recognition}
\label{sec:related_work:lp_recognition}

In~\citep{silva2017realtime}, a YOLO-based model was proposed to simultaneously detect and recognize all characters within a cropped \gls*{lp}.
While impressive \gls*{fps} rates (i.e., $448$~\gls*{fps} on a high-end \acrshort*{gpu}) were attained in experiments carried out in the \ssig dataset~\citep{goncalves2016benchmark}, less than $65$\% of the \glspl*{lp} were correctly recognized. According to the authors, the accuracy bottleneck of their approach was letter recognition since the training set of characters was highly unbalanced (in particular, letters).
Silva \& Jung~\cite{silva2018license,silva2020realtime} retrained that model with an enlarged training set composed of real and artificially generated images using font-types similar to the \glspl*{lp} of certain regions.
In this way, the retrained network became much more robust for the detection and classification of real characters, outperforming previous works and commercial systems in three public~datasets.

Li et al.~\cite{li2018reading} proposed to perform character recognition as a sequence labeling problem, also without the character-level segmentation.
Sequential features were first extracted from the entire \gls*{lp}~patch using a \gls*{cnn} in a sliding window manner. 
Then, \glspl*{brnn} with \gls*{lstm} were applied to label the sequential features.
Lastly, \gls*{ctc} was employed for sequence decoding.
The results showed that this method attained better recognition rates than the baselines. Nonetheless, only \minor{\glspl*{lp} from the Taiwan region} were used in their experiments and the execution time was not~reported.

Dong et al.~\cite{dong2017cnnbased} claimed that the method proposed in~\citep{li2018reading} is very fragile to distortions caused by viewpoint change and therefore is not suitable for \gls*{lp} recognition in the wild. Thus, an \gls*{lp} rectification step is employed first in their approach. Afterward, a \gls*{cnn} was trained to recognize Chinese characters, while a shared-weight \gls*{cnn} recognizer was used for digits and English letters, making full use of the limited training data. The recognition rate attained on a private dataset with \minor{\glspl*{lp} from mainland China} was~$89.05$\%.
The authors did not report the execution time of this particular~stage.

Zhuang et al.~\cite{zhuang2018towards} proposed a semantic segmentation technique followed by a character count refinement module to recognize the characters of an \gls*{lp}.
For semantic segmentation, they simplified the DeepLabV$2$ (ResNet-$101$) model by removing the multi-scaling process, increasing computational efficiency.
Then, the character areas were generated through \gls*{cca}.
Finally, Inception-v$3$ and AlexNet were adopted as the character classification and character counting models, respectively.
The authors claimed that both an outstanding recognition performance and a high computational efficiency were attained. Nevertheless, they assumed that \gls*{lp} detection is easily accomplished and used cropped patches containing only the \gls*{lp} with almost no background (i.e., the ground truth) as input.
Furthermore, their system is not able to process images in real time, especially when considering the time required for the \gls*{lp}~detection stage, which is often more time-consuming than the recognition one.

Some papers focus on deblurring the \glspl*{lp}, which is very useful for \gls*{lp} recognition.
Lu et al.~\cite{lu2016robust} proposed a scheme based on sparse representation to identify the blur kernel, while Svoboda et al.~\cite{svoboda2016cnn} employed a text deblurring \gls*{cnn} for reconstruction of blurred \glspl*{lp}.
Despite achieving exceptional qualitative results, the additional computational cost of a deblurring stage usually is prohibitive for real-time \gls*{alpr}~applications.

\subsection{Final Remarks}
\label{sec:related_work:final_remarks}

The approaches developed for \gls*{alpr} are still limited in various ways.
Many authors only addressed part of the \gls*{alpr} pipeline, e.g., \gls*{lp} detection~\citep{kurpiel2017convolutional,yepez2018improved,xie2018new} or character/\gls*{lp} recognition~\citep{menotti2014vehicle,yang2018chinese,zhuang2018towards}, or performed their experiments on private datasets~\citep{bulan2017segmentation,dong2017cnnbased,yang2018chinese}, making it difficult to accurately evaluate the presented methods.
Note that works focused on a single stage \emph{\textbf{do not consider}} localization errors (i.e., correct but not so accurate detections) in earlier~stages~\cite{spanhel2017holistic,zhuang2018towards}.
Such errors directly affect the recognition results.
As an example,
Gonçalves et al.~\cite{goncalves2018realtime} improved their results by $20$\% by skipping the \gls*{lp} detection stage, that is, by feeding the \glspl*{lp} manually cropped into their recognition~network.

In this work, the proposed end-to-end system is evaluated in eight public datasets that present a great variety in the way they were collected, with images of various types of vehicles (including motorcycles) and numerous \gls*{lp} layouts.
It should be noted that, in most of the works in the literature, no more than three datasets were used in the experiments~(e.g.,~\cite{laroca2018robust,li2018reading,kessentini2019twostage,zhuang2018towards}).
In addition, despite the fact that motorcycles are one of the most popular transportation means in metropolitan areas~\citep{hsu2016comparison}, motorcycle images have not been used in the assessment of most \gls*{alpr} systems in the~literature~\cite{goncalves2018realtime,silva2020realtime}.

Most of the approaches are not capable of recognizing \glspl*{lp} in real time (i.e., $30$~\gls*{fps})~\citep{li2018toward,silva2018license,zhuang2018towards}, even running on high-end GPUs, making it impossible for them to be applied in some real-world applications \minor{(especially considering that the purchase of high-end setups is not practicable for various commercial applications~\cite{castro2020license})}.
Furthermore, several authors do not report the execution time of the proposed methods or report the time required only for a specific stage~\citep{dong2017cnnbased,yang2018chinese,li2018reading}, making it difficult an accurate analysis of their speed/accuracy trade-off, as well as their applicability. 
In this sense, we explore different YOLO models at each stage, carefully optimizing and combining them to achieve the best speed/accuracy trade-off.
In our experiments, both the accuracy and execution time are reported to enable fair comparisons in future~works.

It is important to highlight that although outstanding results in terms of \gls*{map} have been achieved with other object detectors such as \gls*{ssd}~\citep{liu2016ssd} and RetinaNet~\citep{lin2017focal}, in this work we adapt YOLO since it focuses on an \textit{extreme} speed/accuracy trade-off~\citep{lin2017focal}, which is essential in intelligent transportation systems~\cite{xing2019driver}, being able to process more than twice as many \gls*{fps} as other detectors while still achieving competitive results~\citep{redmon2017yolo9000,redmon2018yolov3}.

\begin{figure*}[!htb]
    \centering
    \, \includegraphics[width=0.985\textwidth]{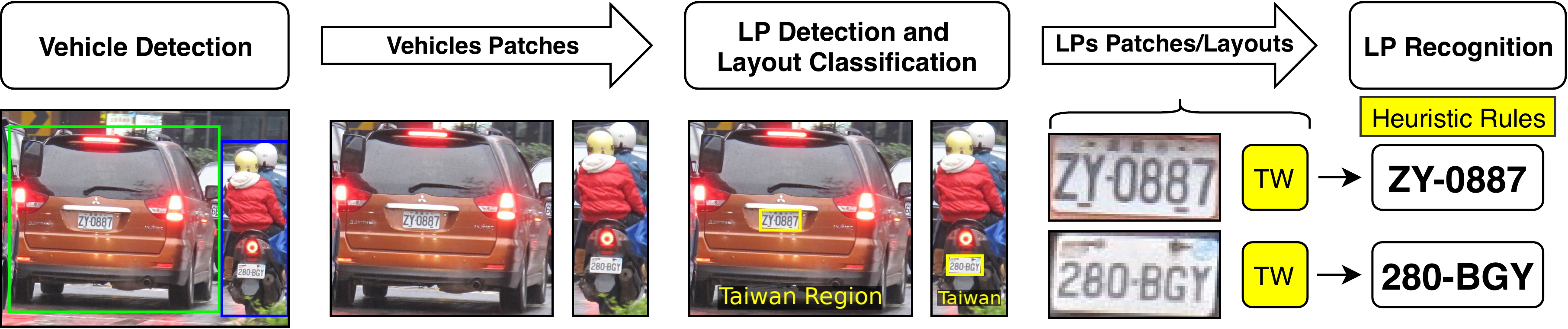}
    
    \vspace{-1.5mm}
    
    \caption{The pipeline of the proposed \gls*{alpr} system.
    First, all vehicles are detected in the input image.
    Then, in a single stage, the \gls*{lp} of each vehicle is detected and its layout is classified (in the example above, \minor{the vehicles/\glspl*{lp} are from the Taiwan region}).
    Finally, all characters of each \gls*{lp} are recognized simultaneously, with heuristic rules being applied to adapt the results according to the predicted layout~class. 
    %
    }
    \label{fig:alpr_pipeline}
    \vspace{-1mm}
\end{figure*}

\minor{
Although YOLO has already been employed in the \gls*{alpr} context, previous works present several limitations (as detailed in sections~\ref{sec:related_work:lp_detection} and~\ref{sec:related_work:lp_recognition}), with the authors commonly overlooking many factors that may limit the accuracy or speed (or even both) achieved by YOLO-based models, such as the dimensions of the input images, number of network layers, filters and anchors, and/or using data augmentation strategies that can actually impair the network~learning.
These factors have not been discussed sufficiently in the~literature.
}

We consider \gls*{lp} recognition as the current bottleneck of \gls*{alpr} systems since (i)~impressive \gls*{lp} detection results have been reported in recent works~\citep{laroca2018robust,xie2018new,alshemarry2018ensemble}, both in terms of recall rate and execution time; (ii)~\gls*{ocr} approaches must work as close as possible to the optimality (i.e.,~$100$\% of character recognition rate) in the \gls*{alpr} context, as a single mistake may imply in incorrect identification of the vehicle~\citep{goncalves2016license}.
Thus, in this work, we propose a unified approach for \textit{\gls*{lp} detection and layout classification} in order to improve the recognition results using heuristic rules.
Additionally, we design and apply data augmentation techniques to simulate \glspl*{lp} of other layouts and also to generate \gls*{lp} images with characters that have few instances in the training set.
Hence, unlike~\citep{silva2017realtime,yang2018chinese}, we avoid errors in the recognition stage due to highly unbalanced training sets of \gls*{lp}~characters.
\section{Proposed ALPR System}
\label{sec:proposed}

The nature of traffic images might be very problematic to LP detection approaches that work directly on the frames (i.e., without vehicle detection) since (i)~there are many textual blocks that can be confused with \glspl*{lp} such as traffic signs and phone numbers on storefronts, and (ii)~\glspl*{lp} might occupy very small portions of the original~image \colored{and object detectors commonly struggle to detect small objects~\cite{redmon2016yolo,liu2019deep,silva2020realtime}}.
Therefore, we propose to first locate the vehicles in the input image and then detect their respective \glspl*{lp} in the vehicle patches.
Afterward, we detect and recognize all characters simultaneously by feeding the entire \gls*{lp} patch into the network.
In this way, we do not need to deal with the character segmentation~task.

Although some approaches with such characteristics (i.e., containing a vehicle detection stage prior to \gls*{lp} detection and/or avoiding character segmentation) have already been proposed in the literature, none of them presented robustness for different \gls*{lp} layouts in both accuracy and processing time.
In~\citep{silva2017realtime} and~\citep{goncalves2018realtime}, for instance, the authors designed real-time \gls*{alpr} systems able to process more than $50$~\gls*{fps} on high-end \acrshortpl*{gpu}, however, both systems were evaluated only on \glspl*{lp} from a single country and presented poor recognition rates in at least one dataset in which they were evaluated.
On the other hand, outstanding results were achieved on different scenarios in some recent works~\citep{li2018toward,li2018reading,silva2018license}, however, the methods presented in these works are computationally expensive and cannot be applied in real time.
This makes them unsuitable for use in many real-world~applications, \minor{especially those where multiple vehicles can appear on the scene at the same time~\cite{goncalves2018realtime}}.

In order to develop an \gls*{alpr} system that is robust for different \gls*{lp} layouts, we propose a \emph{\textbf{layout classification}} stage after \gls*{lp} detection.
However, instead of performing both stages separately, we merge the \gls*{lp} detection and layout classification tasks by training an object detection network that outputs a distinct class for each \gls*{lp} layout. In this way, with almost no additional cost, we employ layout-specific approaches for \gls*{lp} recognition in cases where the \gls*{lp} and its layout are predicted with a confidence value above a predefined threshold.
For example, all Brazilian \glspl*{lp} have seven characters: three letters and four digits (in that order), and thus a post-processing method is applied to avoid errors in characters that are often misclassified, such as `B'~and~`$8$', `G'~and~`$6$', `I'~and~`$1$', among others. In cases where the \gls*{lp} and its layout are detected with confidence below the predefined threshold, a generic approach is~applied.
To the best of our knowledge, this is the first time a layout classification stage is proposed to improve the recognition results.

\colored{
It is worth noting that although the \gls*{lp} layout is known \textit{a priori} in some real-world applications, there are various regions/countries around the world where multiple \gls*{lp} layouts coexist.
As an example, Mercosur countries (Argentina, Brazil, Paraguay and Uruguay) are adopting a new standard of \glspl*{lp} for newly purchased vehicles, inspired by the integrated system adopted several years ago by European Union countries~\cite{mercopress2014mercosur}.
As changing to the new \gls*{lp} layout is not free of charge~\cite{riotimes2020mercosur_a} and is not mandatory for used vehicles~\cite{riotimes2020mercosur_b}, the old and new layouts will coexist for many years in these countries.
In fact, such a situation will occur in any country/region that adopts a new \gls*{lp} layout without drivers being required to update their current \glspl*{lp}, as occurred in most European Union countries in the past.
Hence, in such cases, an \gls*{alpr} system capable of classifying the \gls*{lp} layout is \emph{\textbf{essential}} to avoid errors in the number of predicted characters to be considered and also in similar letters and digits, since the number of characters and/or the positions for letters and digits often differ in the old and new \gls*{lp} layouts (e.g., Argentine `old' \glspl*{lp} consist of exactly 3~letters and 3~digits, whereas the initial pattern adopted in Argentina for Mercosur \glspl*{lp} consists of 2~letters, 3~digits and 2~letters, in that~order).} 

\minor{In this context, although layout-dependent factors can be addressed by developing a tailored \gls*{alpr} system for the specific subset of \gls*{lp} layouts that coexist in a given region, such systems must be verified/modified if a new LP layout is adopted in that region (or if authorities want to start recognizing \glspl*{lp} from neighboring countries) since some previously used strategies may no longer be applicable.
On the other hand, for the proposed approach to work for additional \gls*{lp} layouts, we only need to retrain our network for \gls*{lp} detection and layout classification with images of the new \gls*{lp} layout (in addition to images of the known layouts) and adjust the expected pattern (i.e., the number of characters and fixed positions of digits and letters) in a configuration file.
In other words, layout classification (along with heuristic rules) enables the proposed \gls*{alpr} system to be easily adjusted to work for additional/different LP~layouts.
}

As great advances in object detection have been achieved using YOLO-inspired models~\citep{ning2017spatially,wu2017squeeze,tripathi2017lcdet,severo2018benchmark}, we decided to specialize it for \gls*{alpr}. 
We use specific models for each stage. Thus, we can tune the parameters separately in order to improve the performance of each task.
The models adapted are YOLOv2~\citep{redmon2017yolo9000}, Fast-YOLOv2 and CR-NET~\citep{silva2017realtime}, which is an architecture inspired by YOLO for character detection and recognition.
We explored several data augmentation techniques and performed modifications to each network (e.g., changes in the input size, number of filters, layers and anchors) to achieve the best \emph{\textbf{speed/accuracy trade-off}} at each~stage. 

In this work, unlike~\cite{panahi2017accurate,laroca2018robust,zhuang2018towards}, for each stage, we train a single network on images from several datasets (described in Section~\ref{sec:experiments:datasets}) to make our networks robust for distinct \gls*{alpr} applications or scenarios with considerably less manual effort since their parameters are adjusted only once for all~datasets.

This remainder of this section describes the proposed approach and it is divided into three subsections, one for each stage of our end-to-end \gls*{alpr} system: (i)~vehicle detection, (ii)~\gls*{lp} detection and layout classification and (iii)~\gls*{lp} recognition.
Fig.~\ref{fig:alpr_pipeline} illustrates the system pipeline, explained throughout this~section.

\subsection{Vehicle Detection}
\label{sec:proposed:vehicle_detection}

In this stage, we explored the following models: Fast-YOLOv2, YOLOv2~\citep{redmon2017yolo9000}, Fast-YOLOv3 and YOLOv3~\citep{redmon2018yolov3}.
Although the Fast-YOLO variants correctly located the vehicles in most cases, they failed in challenging scenarios such as images in which one or more vehicles are partially occluded or appear in the background.
On the other hand, impressive results (i.e., F-measure rates above $98$\% in the validation set\footnote{The division of the images of each dataset into training, test and validation sets is detailed in Section~\ref{sec:experiments:evaluation_protocol}.}) were obtained with both YOLOv2 and YOLOv3, which successfully detected vehicles even in those cases where the smaller models failed.
As the computational cost is one of our main concerns and YOLOv3 is much more complex than its predecessor, \minor{we further improve the YOLOv2 model for vehicle~detection}.

First, we changed the network input size from $416\times416$ to $448\times288$ pixels since the images used as input to \gls*{alpr} systems generally have a width greater than height.
Hence, our network processes less distorted images and performs faster, as the new input size is $25$\% smaller than the original.
The new dimensions were chosen based on speed/accuracy assessments with different input sizes (from $448\times288$ to $832\times576$ pixels). 
Then, we recalculate the anchor boxes for the new input size as well as for the datasets employed in our experiments using the k-means clustering algorithm.
Finally, we reduced the number of filters in the last convolutional layer to match the number of classes.
YOLOv2 uses $A$~anchor boxes to predict bounding boxes (we use $A$~=~$5$), each with four coordinates $(x, y, w, h)$, confidence and $C$ class probabilities~\citep{redmon2017yolo9000}, so the number of filters is given by 
\begin{equation} 
\label{eq:filters}
\textit{filters} = (C + 5) \times A \, .
\end{equation} 

\noindent As we intend to detect cars and motorcycles (two classes), the number of filters in the last convolutional layer must be $35$~(($2+5$) $\times~5$). According to preliminary experiments, the results were better when using two classes instead of just one regarding both types of~vehicles. 

The modified YOLOv2 architecture for vehicle detection is shown in Table~\ref{tab:yolov2_vehicle_detection}.
We exploit various data augmentation strategies, such as flipping, rescaling and shearing, to train our network. Thus, we prevent overfitting by creating many other images with different characteristics from a single labeled one.

\begin{table}[!htb]
\centering
\caption{The YOLOv2 architecture, modified for vehicle detection. The input size was changed from $416\times416$ to $448\times288$ pixels and the number of filters in the last layer was reduced from $425$ to $35$.}
\label{tab:yolov2_vehicle_detection}
\vspace{1mm}
\resizebox{0.85\columnwidth}{!}{ 
\begin{tabular}{@{}cccccc@{}}
\toprule
\textbf{\#} & \textbf{Layer} & \textbf{Filters} & \textbf{Size} & \textbf{Input} & \textbf{Output} \\ \midrule
$0$ & conv & $32$ & $3 \times 3 / 1$ & $448 \times 288 \times 3$ & $448 \times 288 \times 32$ \\
$1$ & max &  & $2 \times 2 / 2$ & $448 \times 288 \times 32$ & $224 \times 144 \times 32$ \\
$2$ & conv & $64$ & $3 \times 3 / 1$ & $224 \times 144 \times 32$ & $224 \times 144 \times 64$ \\
$3$ & max &  & $2 \times 2 / 2$ & $224 \times 144 \times 64$ & $112 \times 72 \times 64$ \\
$4$ & conv & $128$ & $3 \times 3 / 1$ & $112 \times 72 \times 64$ & $112 \times 72 \times 128$ \\
$5$ & conv & $64$ & $1 \times 1 / 1$ & $112 \times 72 \times 128$ & $112 \times 72 \times 64$ \\
$6$ & conv & $128$ & $3 \times 3 / 1$ & $112 \times 72 \times 64$ & $112 \times 72 \times 128$ \\
$7$ & max &  & $2 \times 2 / 2$ & $112 \times 72 \times 128$ & $56 \times 36 \times 128$ \\
$8$ & conv & $256$ & $3 \times 3 / 1$ & $56\times 36 \times 128$ & $56 \times 36 \times 256$ \\
$9$ & conv & $128$ & $1 \times 1 / 1$ & $56\times 36 \times 256$ & $56 \times 36 \times 128$ \\
$10$ & conv & $256$ & $3 \times 3 / 1$ & $56\times 36 \times 128$ & $56 \times 36 \times 256$ \\
$11$ & max &  & $2 \times 2 / 2$ & $56 \times 36 \times 256$ & $28 \times 18 \times 256$ \\
$12$ & conv & $512$ & $3 \times 3 / 1$ & $28 \times 18 \times 256$ & $28 \times 18 \times 512$ \\
$13$ & conv & $256$ & $1 \times 1 / 1$ & $28 \times 18 \times 512$ & $28 \times 18 \times 256$ \\
$14$ & conv & $512$ & $3 \times 3 / 1$ & $28 \times 18 \times 256$ & $28 \times 18 \times 512$ \\ 
$15$ & conv & $256$ & $1 \times 1 / 1$ & $28 \times 18 \times 512$ & $28 \times 18 \times 256$ \\
$16$ & conv & $512$ & $3 \times 3 / 1$ & $28 \times 18 \times 256$ & $28 \times 18 \times 512$ \\
$17$ & max &  & $2 \times 2 / 2$ & $28 \times 18 \times 512$ & $14 \times 9 \times 512$ \\
$18$ & conv & $1024$ & $3 \times 3 / 1$ & $14 \times 9 \times 512$ & $14 \times 9 \times 1024$ \\
$19$ & conv & $512$ & $1 \times 1 / 1$ & $14 \times 9 \times 1024$ & $14 \times 9 \times 512$ \\
$20$ & conv & $1024$ & $3 \times 3 / 1$ & $14 \times 9 \times 512$ & $14 \times 9 \times 1024$ \\
$21$ & conv & $512$ & $1 \times 1 / 1$ & $14 \times 9 \times 1024$ & $14 \times 9 \times 512$ \\
$22$ & conv & $1024$ & $3 \times 3 / 1$ & $14 \times 9 \times 512$ & $14 \times 9 \times 1024$ \\
$23$ & conv & $1024$ & $3 \times 3 / 1$ & $14 \times 9 \times 1024$ & $14 \times 9 \times 1024$ \\
$24$ & conv & $1024$ & $3 \times 3 / 1$ & $14 \times 9 \times 1024$ & $14 \times 9 \times 1024$ \\
$25$ & route [16] & & & & \\
$26$ & reorg & & \phantom{aaaa} $/ 2$ & $28 \times 18 \times 512$ & $14 \times 9 \times 2048$ \\
$27$ & route [26, 24] & & & & \\
$28$ & conv & $1024$ & $3 \times 3 / 1$ & $14 \times 9 \times 3072$ & $14 \times 9 \times 1024$ \\
$29$ & conv & $35$ & $1 \times 1 / 1$ & $14 \times 9 \times 1024$ & $14 \times 9 \times 35$ \\
$30$ & detection &  &  &  &  \\ \bottomrule
\end{tabular}
}  
\end{table}

Silva \& Jung~\cite{silva2018license} slightly modified their pipeline by directly applying their \gls*{lp} detector (i.e., skipping the vehicle detection stage) when dealing with images in which the vehicles are very close to the camera, as their detector failed in several of those cases. 
We believe this is not the best way to handle the problem.
Instead, we do not skip the vehicle detection stage even when only a small part of the vehicle is visible. 
The entire image is labeled as ground truth in cases where the vehicles are very close to the camera. Therefore, our network also learns to select the \gls*{roi} in such~cases.

In the validation set, we evaluate several confidence thresholds to detect as many vehicles as possible while maintaining a low \gls*{fp} rate.
Furthermore, we apply a \gls*{nms} algorithm to eliminate redundant detections (those with \gls*{iou} $\ge0.25$) since the same vehicle might be detected more than once by the network. 
A negative recognition result is given in cases where no vehicle is~found.

\subsection{License Plate Detection and Layout Classification}
\label{sec:proposed:lp_detection}

\begin{figure}[!htb]
    \centering
    
    \includegraphics[width=0.95\columnwidth]{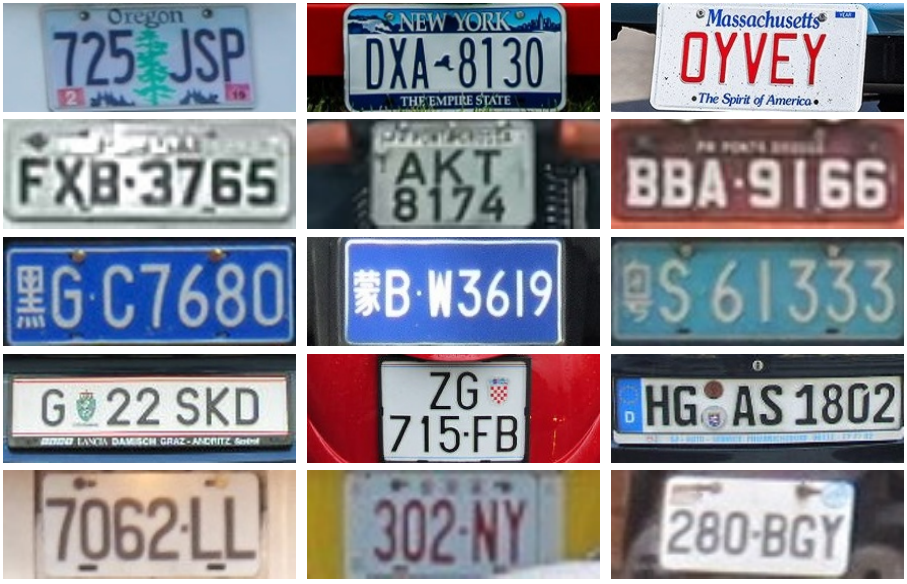}
    
    \vspace{-1.5mm}
    
    \caption{Examples of \glspl*{lp} of different layouts and classes~(from top to bottom: American, Brazilian, Chinese, European and Taiwanese). Observe the wide variety in different ways on different \gls*{lp} layouts.}
    \label{fig:lp_detection_layout_samples}

\end{figure}

In this work, we detect the \gls*{lp} and simultaneously classify its layout into one of the following classes:
\textit{American}, \textit{Brazilian}, \textit{Chinese}~\minor{(\glspl*{lp} of vehicles registered in mainland China)}, \textit{European} or \textit{Taiwanese}~\minor{(\glspl*{lp} of vehicles registered in the Taiwan region)}. 
These classes were defined based on the public datasets found in the literature~\citep{caltech,englishlpd,ucsd,zhou2012principal,hsu2013application,openalpreu,goncalves2016benchmark,laroca2018robust} and also because there are many \gls*{alpr} systems designed primarily for \glspl*{lp} of one of those regions~\citep{hsu2013application,silva2017realtime,yang2018chinese}. 
It is worth noting that (i)~among \glspl*{lp} with different layouts (which may belong to the same class/region) there is a wide variety in many factors, for example, in the aspect ratio, colors, symbols, position of the characters, number of characters, among others; (ii)~we consider \glspl*{lp} from different jurisdictions in the United States as a single class; the same is done for \glspl*{lp} from European countries. 
\glspl*{lp} from the same country or region may look quite different, but still share many characteristics in common.
Such common features can be exploited to improve \gls*{lp}~recognition. In Fig.~\ref{fig:lp_detection_layout_samples}, we show examples of \glspl*{lp} of different layouts and~classes.

Looking for an efficient \gls*{alpr} system, in this stage we performed experiments with the Fast-YOLOv2 and Fast-YOLOv3 models. In the validation set, Fast-YOLOv2 obtained slightly better results than its successor.
This is due to the fact that YOLOv3 and Fast-YOLOv3 have relatively high performance on small objects (which is not the case since we first detect the vehicles), but comparatively worse performance on medium and larger size objects~\citep{redmon2018yolov3}. Accordingly, here we modified the Fast-YOLOv2~model to adapt it to our application and to achieve even better results.

First, we changed the kernel size of the next-to-last convolutional layer from $3\times3$ to $1\times1$.
Then, we added a $3\times3$ convolutional layer with twice the filters of that layer.
In this way, the network reached better results (F-measure~$\approx1$\% higher, from $97.97$\% to $99.00$\%) almost without increasing the number of \gls*{flop} required, i.e., from $5.35$ to $5.53$~\gls*{bflop}, as alternating $1\times1$ convolutional layers between $3\times3$ convolutions reduce the feature space from preceding layers~\citep{redmon2016yolo,redmon2017yolo9000}.
Finally, we recalculate the anchors for our data and make adjustments to the number of filters in the last layer. The modified architecture is shown in Table~\ref{tab:lp_detection_fastyolov2}.

\begin{table}[!htb]
\setlength{\tabcolsep}{7pt}
\centering
\caption{Fast-YOLOv2 modified for \gls*{lp} detection and layout classification. First, we reduced the kernel size of layer~\#$13$ from $3\times3$ to $1\times1$, and added layer~\#$14$. Then, we reduced the number of filters in layer \#$15$ from $425$~to~$50$, as we use $5$ anchor boxes to detect $5$ classes (see Equation~\ref{eq:filters}).}
\label{tab:lp_detection_fastyolov2}
\vspace{1mm}
\resizebox{0.95\columnwidth}{!}{ 
\begin{tabular}{@{}ccccccc@{}}
\toprule
\textbf{\#} & \textbf{Layer} & \textbf{Filters} & \textbf{Size} & \textbf{Input} & \textbf{Output} & \textbf{\gls*{bflop}} \\ \midrule
$0$ & conv & $16$ & $3 \times 3 / 1$ & $416 \times 416 \times 3$ & $416 \times 416 \times 16$ & $0.150$ \\
$1$ & max &  & $2 \times 2 / 2$ & $416 \times 416 \times 16$ & $208 \times 208 \times 16$ & $0.003$ \\
$2$ & conv & $32$ & $3 \times 3 / 1$ & $208 \times 208 \times 16$ & $208 \times 208 \times 32$ & $0.399$ \\
$3$ & max &  & $2 \times 2 / 2$ & $208 \times 208 \times 32$ & $104 \times 104 \times 32$ & $0.001$ \\
$4$ & conv & $64$ & $3 \times 3 / 1$ & $104 \times 104 \times 32$ & $104 \times 104 \times 64$ & $0.399$ \\
$5$ & max &  & $2 \times 2 / 2$ & $104 \times 104 \times 64$ & $52 \times 52 \times 64$ & $0.001$ \\
$6$ & conv & $128$ & $3 \times 3 / 1$ & $52\times 52 \times 64$ & $52 \times 52 \times 128$ & $0.399$ \\
$7$ & max &  & $2 \times 2 / 2$ & $52 \times 52 \times 128$ & $26 \times 26 \times 128$ & $0.000$ \\ 
$8$ & conv & $256$ & $3 \times 3 / 1$ & $26 \times 26 \times 128$ & $26 \times 26 \times 256$ & $0.399$ \\
$9$ & max &  & $2 \times 2 / 2$ & $26 \times 26 \times 256$ & $13 \times 13 \times 256$ & $0.000$ \\
$10$ & conv & $512$ & $3 \times 3 / 1$ & $13 \times 13 \times 256$ & $13 \times 13 \times 512$ & $0.399$ \\
$11$ & max &  & $2 \times 2 / 1$ & $13 \times 13 \times 512$ & $13 \times 13 \times 512$ & $0.000$ \\
$12$ & conv & $1024$ & $3 \times 3 / 1$ & $13 \times 13 \times 512$ & $13 \times 13 \times 1024$ & $1.595$ \\
$13$ & conv & $512$ & $1 \times 1 / 1$ & $13 \times 13 \times 1024$ & $13 \times 13 \times 512$ & $0.177$ \\
$14$ & conv & $1024$ & $3 \times 3 / 1$ & $13 \times 13 \times 512$ & $13 \times 13 \times 1024$ & $1.595$ \\
$15$ & conv & $50$ & $1 \times 1 / 1$ & $13 \times 13 \times 1024$ & $13 \times 13 \times 50$ & $0.017$ \\
$16$ & detection &  &  &  &  \\ \bottomrule
\end{tabular} \,}
\end{table}

In Table~\ref{tab:lp_detection_fastyolov2}, we also list the number of \gls*{flop} required in each layer to highlight how small the modified network is compared to others, e.g., YOLOv2 and YOLOv3. For this task, our network requires $5.53$ \gls*{bflop} while YOLOv2 and YOLOv3 require $29.35$ and $66.32$ \gls*{bflop}, respectively.
It is noteworthy that we only need to increase the number of filters (following Equation~\ref{eq:filters}) in the last convolutional layer so that the network can detect/classify additional \gls*{lp}~layouts.

For \gls*{lp} detection and layout classification, we also use data augmentation strategies to generate many other images from a single labeled one.
However, horizontal flipping is not performed at this stage, as the network leverages information such as the position of the characters and symbols on the \gls*{lp} to predict its layout (besides the aspect ratio, colors, and other~characteristics).

Only the detection with the highest confidence value is considered in cases where more than one \gls*{lp} is predicted, as each vehicle has only one \gls*{lp}.
Then, we classify as `undefined~layout' every \gls*{lp} that has its position and class predicted with a confidence value below $0.75$, regardless of which class the network predicted (note that such \glspl*{lp} are not rejected, instead, a generic approach is used in the recognition stage).
This threshold was chosen based on experiments performed in the validation set, in which approximately $92$\% of the \glspl*{lp} were predicted with a confidence value above $0.75$. 
In each of these cases, the \gls*{lp} layout was correctly classified.
A negative result is given in cases where no \gls*{lp} is predicted by the network.

\subsection{License Plate Recognition}
\label{sec:proposed:lp_recognition}

Once the \gls*{lp} has been detected and its layout classified, we employ CR-NET~\cite{silva2017realtime} for \gls*{lp} recognition~(i.e.,~all characters are recognized simultaneously by feeding the entire \gls*{lp} patch into the network). CR-NET is a model that consists of the first eleven layers of YOLO and four other convolutional layers added to improve nonlinearity.
This model was chosen for two main reasons.
First, it was capable of detecting and recognizing \gls*{lp} characters at $448$ \gls*{fps} in~\citep{silva2017realtime}
Second, very recently, it yielded the best recognition results in the context of image-based automatic meter reading~\citep{laroca2019convolutional}, outperforming two segmentation-free approaches based on deep~learning.

\begin{figure}[!htb]
    \centering
    
    \includegraphics[width=0.85\columnwidth]{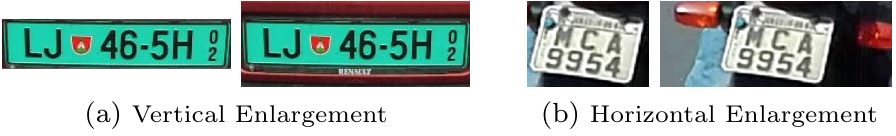}
    
    \vspace{-2mm}
    
    \caption[Two illustrations of enlargement of the \glspl*{lp} detected in the \gls*{lp} detection stage]{Two illustrations of enlargement of the \glspl*{lp} detected in the previous stage. In this way, a single network is trained to recognize \glspl*{lp} of different layouts, regardless of their aspect ratios.}
    \label{fig:enlarge_lps}
\end{figure}

The CR-NET architecture is shown in Table~\ref{tab:cr_net}.
We changed its input size, which was originally defined based on Brazilian \glspl*{lp}, from $240\times80$ to $352\times128$ pixels taking into account the average aspect ratio of the \glspl*{lp} in the datasets used in our experiments, in addition to results obtained in the validation set, where several input sizes were evaluated (e.g., $256\times96$ and $384\times128$ pixels).
As the same model is employed to recognize \glspl*{lp} of various layouts, we enlarge all \gls*{lp} patches (in both the training and testing phases) so that they have aspect ratios~($w/h$) between~$2.5$ and~$3.0$, as shown in Fig.~\ref{fig:enlarge_lps},
considering that the input image has an aspect ratio of~$2.75$.
The network is trained to predict $35$ classes ($0$-$9$, A-Z, where the letter~`O' is detected/recognized jointly with the digit~`0') using the \gls*{lp} patch as well as the class and coordinates of each character as~inputs.

\begin{table}[!htb]
\centering
\caption{The CR-NET model. We increased the input size from $240\times80$ to $352\times128$ pixels. The number of filters in the last convolutional layer (\#$14$) was defined following Equation~\ref{eq:filters} (using~$A$~=~$5$).}
\label{tab:cr_net}
\vspace{1mm}
\resizebox{0.9\columnwidth}{!}{ 
\begin{tabular}{@{}ccccccc@{}}
\toprule
\textbf{\#} & \textbf{Layer} & \textbf{Filters} & \textbf{Size} & \textbf{Input} & \textbf{Output} & \textbf{\gls*{bflop}} \\ \midrule
$0$ & conv & $32$ & $3 \times 3 / 1$ & $352 \times 128 \times 3$ & $352 \times 128 \times 32$ & $0.078$ \\
$1$ & max &  & $2 \times 2 / 2$ & $352 \times 128 \times 32$ & $176 \times 64 \times 32$ & $0.001$ \\
$2$ & conv & $64$ & $3 \times 3 / 1$ & $176 \times 64 \times 32$ & $176 \times 64 \times 64$ & $0.415$\\
$3$ & max &  & $2 \times 2 / 2$ & $176 \times 64 \times 64$ & $88 \times 32 \times 64$ & $0.001$ \\
$4$ & conv & $128$ & $3 \times 3 / 1$ & $88 \times 32 \times 64$ & $88 \times 32 \times 128$ & $0.415$ \\
$5$ & conv & $64$ & $1 \times 1 / 1$ & $88 \times 32 \times 128$ & $88 \times 32 \times 64$ & $0.046$ \\
$6$ & conv & $128$ & $3 \times 3 / 1$ & $88 \times 32 \times 64$ & $88 \times 32 \times 128$ & $0.415$ \\
$7$ & max & & $2 \times 2 / 2$ & $88 \times 32 \times 128$ & $44 \times 16 \times 128$ & $0.000$ \\
$8$ & conv & $256$ & $3 \times 3 / 1$ & $44 \times 16 \times 128$ & $44 \times 16 \times 256$ & $0.415$ \\
$9$ & conv & $128$ & $1 \times 1 / 1$ & $44 \times 16 \times 256$ & $44 \times 16 \times 128$ & $0.046$\\
$10$ & conv & $256$ & $3 \times 3 / 1$ & $44 \times 16 \times 128$ & $44 \times 16 \times 256$ & $0.415$\\
$11$ & conv & $512$ & $3 \times 3 / 1$ & $44 \times 16 \times 256$ & $44 \times 16 \times 512$ & $1.661$ \\
$12$ & conv & $256$ & $1 \times 1 / 1$ & $44 \times 16 \times 512$ & $44 \times 16 \times 256$ & $0.185$ \\
$13$ & conv & $512$ & $3 \times 3 / 1$ & $44 \times 16 \times 256$ & $44 \times 16 \times 512$ & $1.661$ \\
$14$ & conv & $200$ & $1 \times 1 / 1$ & $44 \times 16 \times 512$ & $44 \times 16 \times 200$ & $0.144$ \\
$15$ & detection &  &  &  &  \\ \bottomrule
\end{tabular}
}
\end{table}

It is worth to mention that the first character in Chinese \glspl*{lp} (see Fig.~\ref{fig:lp_detection_layout_samples}) is a Chinese character that represents the province in which the vehicle is affiliated~\citep{liu2018convolutional,yang2018chinese}.
Following~\citep{li2018toward}, our network was not trained/designed to recognize Chinese characters, even though Chinese \glspl*{lp} are used in the experiments. In other words, only digits and English letters are considered. The reason is threefold: (i)~there are less than $400$ images in the \chinese dataset~\citep{zhou2012principal} (only some of them are used for training), which is employed in the experiments, and some provinces are not represented; (ii)~labeling the class of Chinese characters is not a trivial task for non-Chinese people (we manually labeled the position and class of the \gls*{lp} characters in the \chinese dataset); and (iii)~to fairly compare our system with others trained only on digits and English letters.
We remark that in the literature the approaches capable of recognizing Chinese characters, digits and English letters were evaluated, for the most part, on datasets containing only \glspl*{lp} from \minor{mainland China}~\citep{gou2016vehicle,liu2018convolutional,yang2018chinese}.

As the \gls*{lp} layout is classified in the previous stage, we design \textit{heuristic rules} to adapt the results produced by CR-NET according to the predicted class. Based on the datasets employed in this work, we defined the minimum and the maximum number of characters to be considered in \glspl*{lp} of each layout.
Brazilian and Chinese \glspl*{lp} have a fixed number of characters, while American, European and Taiwanese \glspl*{lp} do not (see Table~\ref{tab:lp_recognition_heuristics}).
Initially, we consider all characters predicted with a confidence value above a predefined threshold.
Afterward, as in the vehicle detection stage, an \gls*{nms} algorithm is applied to remove redundant detections.
Finally, if necessary, we discard the characters predicted with lower confidence values or consider others previously discarded (i.e., ignoring the confidence threshold) so that the number of characters considered is within the range defined for the predicted class. 
We consider that the \gls*{lp} has between $4$ and $8$ characters in cases where its layout was classified with a low confidence value (i.e., undefined~layout).

\begin{table}[!htb]
\centering

\caption{The minimum and maximum number of characters to be considered in \glspl*{lp} of each layout class.}
\label{tab:lp_recognition_heuristics}
\vspace{1mm}
\resizebox{0.99\columnwidth}{!}{
\begin{tabular}{@{}cccccc@{}}
\toprule
Characters & American & Brazilian & Chinese & European & Taiwanese \\ \midrule
Minimum & $4$ & $7$ & $6$ & $5$ & $5$ \\
Maximum & $7$ & $7$ & $6$ & $8$ & $6$ \\ \bottomrule
\end{tabular}\,}
\end{table}

Additionally, inspired by Silva \& Jung~\cite{silva2017realtime}, we swap digits and letters on Brazilian and Chinese \glspl*{lp}, as there are fixed positions for digits or letters in those layouts. 
In Brazilian \glspl*{lp}, the first three characters correspond to letters and the last four to digits; while in Chinese \glspl*{lp} the second character is a letter that represents a city in the province in which the vehicle is affiliated. 
This swap approach is not employed for \glspl*{lp} of other layouts since each character position can be occupied by either a letter or a digit in American, European and Taiwanese \glspl*{lp}.
The specific swaps are given by [$1$~$\Rightarrow$~I; $2$~$\Rightarrow$~Z; $4$~$\Rightarrow$~A; $5$~$\Rightarrow$~S; $6$~$\Rightarrow$~G; $7$~$\Rightarrow$~Z; $8$~$\Rightarrow$~B] and [A~$\Rightarrow$~$4$; B~$\Rightarrow$~$8$; D~$\Rightarrow$~$0$; G~$\Rightarrow$~$6$; I~$\Rightarrow$~$1$; J~$\Rightarrow$~$1$; Q~$\Rightarrow$~$0$; S~$\Rightarrow$~$5$; Z~$\Rightarrow$~$7$]. In this way, we avoid errors in characters that are often misclassified.

The \gls*{lp} characters might also be arranged in two rows instead of one. We distinguish such cases based on the predictions of the vehicle type, \gls*{lp} layout, and character coordinates. In our experiments, only two datasets have \glspl*{lp} with the characters arranged in two rows. These datasets were captured in Brazil and Croatia. 
In Brazil, car and motorcycle \glspl*{lp} have the characters arranged in one and two rows, respectively. Thus, we look at the predicted class in the vehicle detection stage in those cases. 
In Croatia, on the other hand, cars might also have \glspl*{lp} with two rows of characters. Therefore, for European \glspl*{lp}, we consider that the characters are arranged in two rows in cases where the bounding boxes of half or more of the predicted characters are located entirely below another character. In our tests, this simple rule was sufficient to distinguish \glspl*{lp} with one and two rows of characters even in cases where the \gls*{lp} is considerably inclined. 
We emphasize that segmentation-free approaches (e.g.,~\cite{bulan2017segmentation,spanhel2017holistic,goncalves2018realtime,goncalves2019multitask}) cannot recognize \glspl*{lp} with two rows of characters, contrarily to YOLO-based approaches, which are better suited to recognize them thanks to YOLO's versatility and ability to learn general component features, regardless of their~positions~\cite{kessentini2019twostage}. 

\begin{figure}[!htb]
    \centering

    \includegraphics[width=0.85\columnwidth]{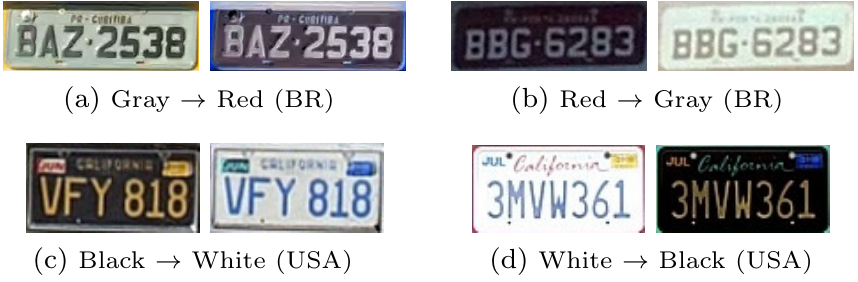} 
        
    \vspace{-2mm}

    \caption{Examples of negative images created to simulate \glspl*{lp} of other layouts.
    In Brazil, private vehicles have gray \glspl*{lp}, while buses, taxis and other transportation vehicles have red \glspl*{lp}. In the United States, old California \glspl*{lp} featured gold characters on a black background. Currently, they have blue characters on a white background.}
    \label{fig:lp_recognition_negative_lps}
\end{figure}

In addition to using the original \gls*{lp} images, we design and apply data augmentation techniques to train the CR-NET model and improve its robustness.
First, we double the number of training samples by creating a negative image of each \gls*{lp}, as we noticed that in some cases negative \glspl*{lp} are very similar to \glspl*{lp} of other layouts. 
This is illustrated with Brazilian and American \glspl*{lp} in Fig.~\ref{fig:lp_recognition_negative_lps}. 
We also generate many other images by randomly rescaling the \gls*{lp} patch and adding a margin to it, simulating more or less accurate detections of the \gls*{lp} in the previous stage.

The datasets for \gls*{alpr} are generally very unbalanced in terms of character classes due to \gls*{lp} allocation policies.
It is well-known that unbalanced data is undesirable for neural network classifiers since the learning of some patterns might be biased.
To address this issue, we permute on the \glspl*{lp} the characters overrepresented in the training set by those underrepresented.
In this way, as in~\cite{goncalves2018realtime}, we are able to create a balanced set of images in which the order and frequency of the characters on the \glspl*{lp} are chosen to uniformly distribute them across the positions. 
We maintain the initial arrangement of letters and digits of each \gls*{lp} so that the network might also learn the positions of letters and digits in certain \gls*{lp} layouts.

\begin{figure}[!htb]
	\centering
	
    \includegraphics[width=0.85\columnwidth]{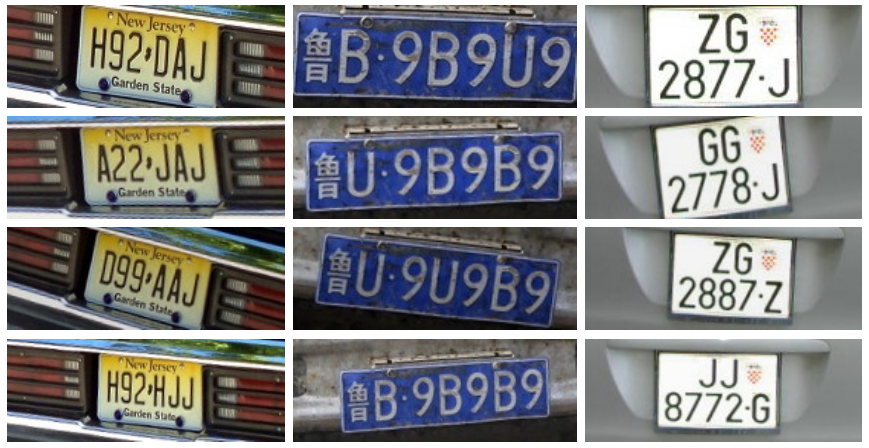}
    
    \vspace{-1.5mm}
	
	\caption{Examples of \gls*{lp} images generated by permuting the characters on the \glspl*{lp}. The images in the first row are the originals and the others were generated~automatically.}
	\label{fig:lp_recognition_data_augmentation}  
\end{figure}

Fig.~\ref{fig:lp_recognition_data_augmentation} shows some artificially generated images by permuting the characters on \glspl*{lp} of different layouts.
We also perform random variations of brightness, rotation and cropping to increase even more the diversity of the generated images.
The parameters were empirically adjusted through visual inspection, i.e., brightness variation of the pixels $[0.85; 1.15]$, rotation angles between \minus$5\degree$~and~$5\degree$ and cropping from \minus$2$\%~to~$8$\% of the \gls*{lp} size.
Once these ranges were established, new images were generated using random values within those ranges for each~parameter.

\section{Experimental Setup}
\label{sec:experiments}

All experiments were performed on a computer with an AMD Ryzen Threadripper $1920$X $3.5$GHz~CPU, $32$ GB of RAM~\minor{(2400 MHz), HDD 7200 RPM}, and an NVIDIA Titan Xp GPU.
The Darknet framework~\citep{darknet} was employed to train and test our networks.
However, we used Alexey Bochkovskiy's version of Darknet~\citep{alexeyab}, which has several improvements over the original, including improved neural network performance by merging two layers into one (convolutional and batch normalization), optimized memory allocation during network resizing, and many other code fixes. For more details on this repository, refer to~\citep{alexeyab}.

We also made use of the Darknet's built-in data augmentation, which creates a number of randomly cropped and resized images with changed colors (hue, saturation, and exposure).
We manually implemented the flip operation only for the vehicle detection stage, as this operation would probably impair the layout classification and the \gls*{lp} recognition tasks.
Similarly, we disabled the color-related data augmentation for the \gls*{lp}~detection and layout~classification stage~(further explained in Section~\ref{sec:results:lp_detection}).

\subsection{Datasets}
\label{sec:experiments:datasets}

The experiments were carried out in \emph{\textbf{eight}} publicly available datasets: \caltech~\citep{caltech}, \englishlpd~\citep{englishlpd}, \stills~\citep{ucsd}, \chinese~\citep{zhou2012principal}, \aolp~\citep{hsu2013application}, \openalpreu~\citep{openalpreu}, \ssig~\citep{goncalves2016benchmark} and \dataset~\citep{laroca2018robust}.
These datasets are often used to evaluate \gls*{alpr} systems, contain multiple \gls*{lp} layouts and were collected under different conditions/scenarios (e.g., with variations in lighting, camera position and settings, and vehicle types). 
An overview of the datasets is presented in Table~\ref{tab:experiments:overview_datasets}.
It is noteworthy that in most of the works in the literature, including some recent ones~\cite{laroca2018robust,li2018reading,kessentini2019twostage,zhuang2018towards}, no more than three datasets were used in the experiments.

\begin{table}[!htb]
\centering
\caption{An overview of the datasets used in our experiments.}
\label{tab:experiments:overview_datasets}
\vspace{1mm}
\resizebox{0.99\columnwidth}{!}{ 
\begin{tabular}{@{}cccccc@{}}
\toprule
\textbf{Dataset} & \textbf{Year} & \textbf{Images} & \textbf{Resolution} & \textbf{LP Layout} & \textbf{\begin{tabular}[c]{@{}c@{}}Evaluation\\ Protocol\end{tabular}} \\ \midrule
\caltech & $1999$ & $126$ & $896\times592$ & American & No \\
\englishlpd & $2003$ & $509$ & $640\times480$ & European & No \\
\stills & $2005$ & $291$ & $640\times480$ & American & Yes \\
\chinese & $2012$ & $411$ & Various & Chinese & No \\
\aolp & $2013$ & $2{,}049$ & Various & Taiwanese & No \\
\openalpreu & $2016$ & $108$ & Various & European & No \\
\ssig & $2016$ & $2{,}000$ & $1{,}920\times1{,}080$ & Brazilian & Yes \\
\dataset & $2018$ & $4{,}500$ & $1{,}920\times1{,}080$ & Brazilian & Yes \\ \bottomrule
\end{tabular}
}
\end{table}

The datasets collected in the United States (i.e., \caltech and \stills) and in Europe (i.e., \englishlpd and \openalpreu) are relatively simple and have certain characteristics in common, for example, most images were captured with a hand-held camera and there is only one vehicle (generally well-centered) in each image.
There are only a few cases in which the \glspl*{lp} are not well aligned.
The \chinese and \aolp datasets, on the other hand, also contain images where the \gls*{lp} is inclined/tilted, as well as images with more than one vehicle, which may be occluded by others.
Lastly, the \ssig and \dataset datasets are composed of high-resolution images, enabling \gls*{lp} recognition from distant vehicles.
In both datasets, there are several frames of each vehicle and, therefore, redundant information may be used to improve the recognition~results.
\colored{Among the eight datasets used in our experiments, we consider the \dataset dataset the most challenging, as three different non-static cameras were used to capture images from different types of vehicles (cars, motorcycles, buses and trucks) with complex backgrounds and under different lighting~conditions~\cite{laroca2018robust}.
Note that both the vehicles and the camera (inside another vehicle) were moving and most \glspl*{lp} occupy a very small region of the~image.}

Most datasets have no annotations or contain labels for a single stage only (e.g., \gls*{lp}~detection), despite the fact that they are often used to train/evaluate algorithms in the \gls*{alpr} context.
Therefore, in all images of these datasets, we manually labeled the position of the vehicles (including those in the background where the \gls*{lp} is also legible), \glspl*{lp} and characters, as well as their classes.

In addition to using the training images of the datasets, we downloaded and labeled more $772$ images from the internet to train all stages of our \gls*{alpr} system. 
This procedure was adopted to eliminate biases from the datasets employed in our experiments. For example, the \caltech and \stills datasets have similar characteristics (e.g., there is one vehicle per image, the vehicle is centered and occupies a large portion of the image, and the resolutions of the images are not high), which are different from those of the other datasets.
Moreover, there are many more examples of Brazilian and Taiwanese \glspl*{lp} in our training data (note that the exact number of images used for training, testing and validation in each dataset is detailed in the next section).
Therefore, we downloaded images containing vehicles with American, Chinese and European \glspl*{lp} so that there are at least $500$ images of \glspl*{lp} of each class/region to train our networks.
Specifically, we downloaded $257$, $341$, and $174$ images containing American, Chinese and European \glspl*{lp},~respectively\footnote{The images were downloaded from \url{www.platesmania.com}. We also made their download links and annotations publicly available.}.

In our experiments, we did not make use of two datasets proposed recently: \aolpe~\citep{hsu2017robust} (an extension of the \aolp dataset) and \gls*{ccpd}~\citep{xu2018towards}.
The former has not yet been made available by the authors, who are collecting more data to make it even more challenging. The latter, although already available, does not provide the position of the vehicles and the characters in its $250{,}000$ images and it would be impractical to label them to train/evaluate our networks (Xu et al.~\cite{xu2018towards} used more than $100{,}000$ images for training in their~experiments).

\subsection{Evaluation Protocol}
\label{sec:experiments:evaluation_protocol}

To evaluate the stages of (i)~vehicle detection and (ii)~\gls*{lp} detection and layout classification, we report the precision and recall rates achieved by our networks. Each metric has its importance since, for system efficiency, all vehicles/\glspl*{lp} must be detected without many false positives.
Note that the precision and recall rates are equal in the \gls*{lp} detection and layout classification stage because we consider only one \gls*{lp} per~vehicle.

We consider as correct only the detections with \gls*{iou} greater than~$0.5$ with the ground truth. This bounding box evaluation, defined in the PASCAL \acrshort*{voc} Challenge~\citep{everingham2010pascalvoc} and employed in previous works~\citep{yuan2017robust,li2018toward,kessentini2019twostage}, is interesting since it penalizes both over- and under-estimated objects.
In the \gls*{lp} detection and layout classification stage, we assess only the predicted bounding box on \glspl*{lp} classified as undefined layout (see Section~\ref{sec:proposed:lp_detection}).
In other words, we consider as correct the predictions when the \gls*{lp} position is correctly predicted but not its layout, as long as the \gls*{lp} (and its layout) has not been predicted with a high confidence value (i.e., below~$0.75$).

In the \gls*{lp} recognition stage, we report the number of correctly recognized \glspl*{lp} divided by the total number of \glspl*{lp} in the test set.
A correctly recognized \gls*{lp} means that all characters on the \gls*{lp} were correctly recognized, as a single character recognized incorrectly may imply in incorrect identification of the vehicle~\cite{goncalves2016license}.

According to Table~\ref{tab:experiments:overview_datasets}, only three of the eight datasets used in this work contain an evaluation protocol (defined by the respective authors) that can be reproduced perfectly: \stills, \ssig and \dataset.
Thus, we split their images into training, validation, and test sets according to their own protocols.
We randomly divided the other five datasets using the protocols employed in previous works, aiming at a fair comparison with them.
In the next paragraph, such protocols (\textit{which we also provide for reproducibility purposes}) are~specified.

We used $80$ images of the \caltech dataset for training and $46$~for testing, as in~\citep{xiang2017license,xiang2019lightweight,zhang2018vehicle}.
Then, we employed $16$ of the $80$ training images for validation (i.e.,~$20$\%). 
The \englishlpd dataset was divided in the same way as in~\citep{panahi2017accurate}, with $80$\% of the images being used for training and the remainder for testing. 
Also in this dataset, $20$\% of the training images were employed for validation.
Regarding the \chinese dataset, we did not find any previous work in which it was split into training/test sets, that is, all its images were used either to train or to test the methods proposed in~\citep{li2018reading,qian2015robust,tian2017license,tian2015twostage}, often jointly with other datasets. 
Thus, we adopted the same protocol of the \ssig and \dataset datasets, in which $40$\% of the images are used for training, $40$\% for testing and $20$\% for validation. 
The \aolp dataset is categorized into three subsets, which represent three major \gls*{alpr} applications: \gls*{ac}, traffic~\gls*{le}, and \gls*{rp}.
As this dataset has been divided in several ways in the literature, we divided each subset into training and test sets with a $2$:$1$ ratio, following~\citep{xie2018new,zhuang2018towards}. 
Then, $20$\% of the training images were employed for validation.
Lastly, all images belonging to the \openalpreu dataset were used for testing in~\citep{masood2017sighthound,silva2018license,silva2020realtime}, while other public or private datasets were employed for training. 
Therefore, we also did not use any image of this dataset for training or validation, only for testing. 
An overview of the number of images used for training, testing and validation in each dataset can be seen in Table~\ref{tab:results:overview_datasets_protocols}.

\begin{table}[!htb]
\centering
\caption{An overview of the number of images used for training, testing and validation in each dataset.}
\label{tab:results:overview_datasets_protocols}
\vspace{1mm}
\resizebox{0.9\columnwidth}{!}{
\begin{tabular}{@{}cccccc@{}}
\toprule
\textbf{Dataset} & \textbf{Training} & \textbf{Validation} & \textbf{Testing} & \textbf{Discarded} & \textbf{Total} \\ \midrule
\caltech & $62$ & $16$ & $46$ & $2$ & $126$\\
\englishlpd & $326$ & $81$ & $102$ & $0$ & $509$ \\
\stills & $181$ & $39$ & $60$ & $11$ & $291$\\
\chinese & $159$ & $79$ & $159$ & $14$ & $411$\\
\aolp & $1{,}093$ & $273$ & $683$ & $0$ & $2{,}049$\\
\openalpreu & $0$ & $0$ & $108$ & $0$ & $108$ \\
\ssig & $789$ & $407$ & $804$ & $0$ & $2{,}000$\\
\dataset & $1{,}800$ & $900$ & $1{,}800$ & $0$ & $4{,}500$ \\ \bottomrule
\end{tabular}
}
\end{table}

\begin{figure}[!htb]
    \centering

    \includegraphics[width=0.95\columnwidth]{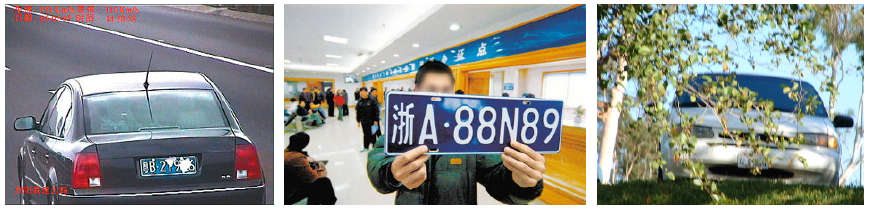} \,
    
    \vspace{-2mm}
    
    \caption{Examples of images discarded in our experiments.}
    \label{fig:results:discarded_images}
\end{figure}

We discarded a few images from the \caltech, \stills, and \chinese datasets\footnote{The list of discarded images can be found at \url{\supplementary}.}.
Although most images in these datasets are reasonable, there are a few exceptions where (i)~it is impossible to recognize the vehicle's \gls*{lp} due to occlusion, lighting or image acquisition problems, etc.; (ii)~the image does not represent real \gls*{alpr} scenarios, for example, a person holding an \gls*{lp}.
Three examples are shown in~Fig.~\ref{fig:results:discarded_images}.
Such images were also discarded in~\citep{masood2017sighthound}.

It is worth noting that we did not discard any image from the test set of the \stills dataset and used the same number of test images in the \caltech dataset. In this way, we can fairly compare our results with those obtained in previous works. 
In fact, we used fewer images from those datasets to train and validate our networks. 
In the \chinese dataset, on the other hand, we first discard the few images with problems and then split the remaining ones using the same protocol as the \ssig and \dataset datasets (i.e.,~$40$/$20$/$40$\% for training, validation and testing, respectively) since, in the literature, a division protocol has not yet been proposed for the \chinese dataset, to the best of our~knowledge.

To avoid an overestimation or bias in the random division of the images into the training, validation and test subsets, we report in each stage the average result of \emph{\textbf{five runs}} of the proposed approach (note that most works in the literature, including recent ones~\citep{laroca2018robust,silva2018license,zhuang2018towards,li2018reading,li2018toward}, report the results achieved in a single run only).
Thus, at each run, the images of the datasets that do not have an evaluation protocol were randomly redistributed into each subset (training/validation/test). 
In the \stills, \ssig and \dataset datasets, we employed the same division (i.e., the one proposed along with the respective dataset) in all runs.

As pointed out in Section~\ref{sec:experiments:datasets}, we manually labeled the vehicles in the background of the images in cases where their \glspl*{lp} are legible. Nevertheless, in the testing phase, we considered only the vehicles/\glspl*{lp} originally labeled in the datasets that have annotations to perform a fair comparison with previous~works.

\section{Results and Discussion}
\label{sec:results}

In this section, we report the experiments carried out to verify the effectiveness of the proposed \gls*{alpr} system.
We first assess the detection stages separately since the regions used in the \gls*{lp} recognition stage are from the detection results, rather than cropped directly from the ground truth.
This is done to provide a realistic evaluation of the entire \gls*{alpr} system, in which well-performed vehicle and \gls*{lp} detections are essential for achieving outstanding recognition results.
Afterward, our system is evaluated in an end-to-end manner and the results achieved are compared with those obtained in previous works and by commercial~systems.

\subsection{Vehicle Detection}
\label{sec:results:vehicle_detection}

In this stage, we employed a confidence threshold of~$0.25$ (defined empirically) to detect as many vehicles as possible, while avoiding high \gls*{fp} rates and, consequently, a higher cost of the proposed \gls*{alpr} system.
The following parameters were used for training the network:~$60$K iterations~(max batches) and learning rate~=~[$10$\textsuperscript{-$3$},~$10$\textsuperscript{-$4$},~$10$\textsuperscript{-$5$}] with steps at $48$K and $54$K~iterations.

\begin{figure}[!htb]
    \centering
    
    \includegraphics[width=0.99\columnwidth]{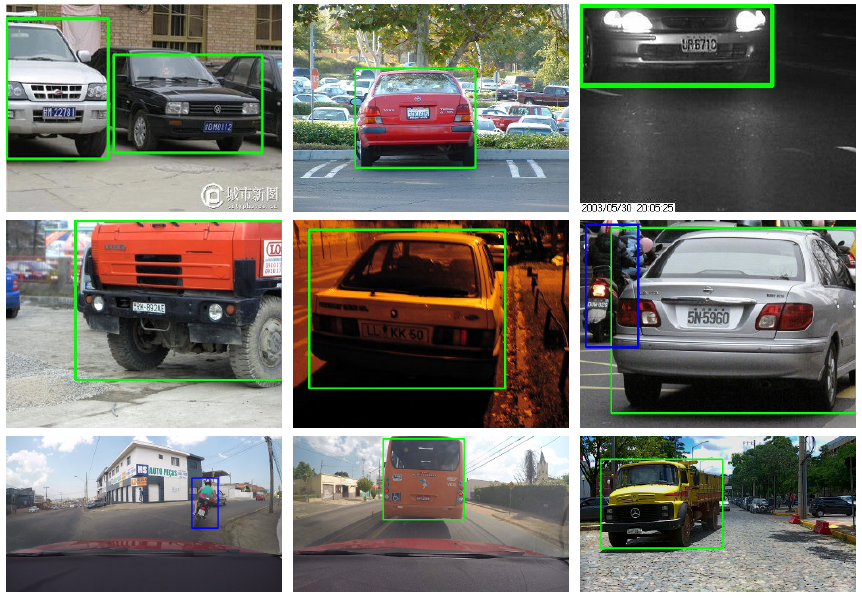} 
    
    \vspace{-2mm}
    
    \caption{Some vehicle detection results achieved in distinct datasets. Observe that vehicles of different types were correctly detected regardless of lighting conditions~(daytime and nighttime), occlusion, camera distance, and other factors.}
    \label{fig:results:veicle_detection_tps}
\end{figure}

The vehicle detection results are presented in Table~\ref{tab:results:vehicle_detection}.
In the average of five runs, our approach achieved a recall rate of $99.92$\% and a precision rate of $98.37$\%. 
It is remarkable that the network was able to correctly detect all vehicles (i.e., recall = $100$\%) in $5$ of the $8$ datasets used in the experiments.
Some detection results are shown in Fig.~\ref{fig:results:veicle_detection_tps}. 
As can be seen, well-located predictions were attained on vehicles of different types and under different conditions.

\begin{table}[!htb]
\centering
\caption{Vehicle detection results achieved across all datasets.}
\label{tab:results:vehicle_detection}
\vspace{1mm}
\resizebox{0.725\columnwidth}{!}{
\begin{tabular}{@{}ccc@{}}
\toprule
\textbf{Dataset} & \textbf{Precision (\%)} & \textbf{Recall (\%)} \\ \midrule
\caltech & $100.00\pm0.00$ & $100.00\pm0.00$ \\
\englishlpd & $99.04\pm0.96$ & $100.00\pm0.00$ \\
\stills & $97.42\pm1.40$ & $100.00\pm0.00$ \\
\chinese & $99.26\pm1.00$ & $99.50\pm0.52$ \\
\aolp & $96.92\pm0.37$ & $99.91\pm0.08$ \\
\openalpreu & $99.27\pm0.76$ & $100.00\pm0.00$ \\
\ssig & $95.47\pm0.62$ & $99.98\pm0.06$ \\
\dataset & $99.57\pm0.07$ & $100.00\pm0.00$ \\ \midrule
\textbf{Average} & $\textbf{98.37}\boldsymbol{\pm}\textbf{0.65}$ & $\textbf{99.92}\boldsymbol{\pm}\textbf{0.08}$ \\ \bottomrule
\end{tabular}
}
\end{table}

To the best of our knowledge, with the exception of the preliminary version of this work~\citep{laroca2018robust}, there is no other work in the \gls*{alpr} context where both cars and motorcycles are detected at this stage.
This is of paramount importance since motorcycles are one of the most popular transportation means in metropolitan areas, especially in Asia~\citep{hsu2016comparison}.
Although motorcycle \glspl*{lp} may be correctly located by \gls*{lp} detection approaches that work directly on the frames, they can be detected with fewer false positives if the motorcycles are detected~first~\citep{hsu2015comparison}. 

\begin{figure}[!htb]
    \centering
    
    \includegraphics[width=0.99\columnwidth]{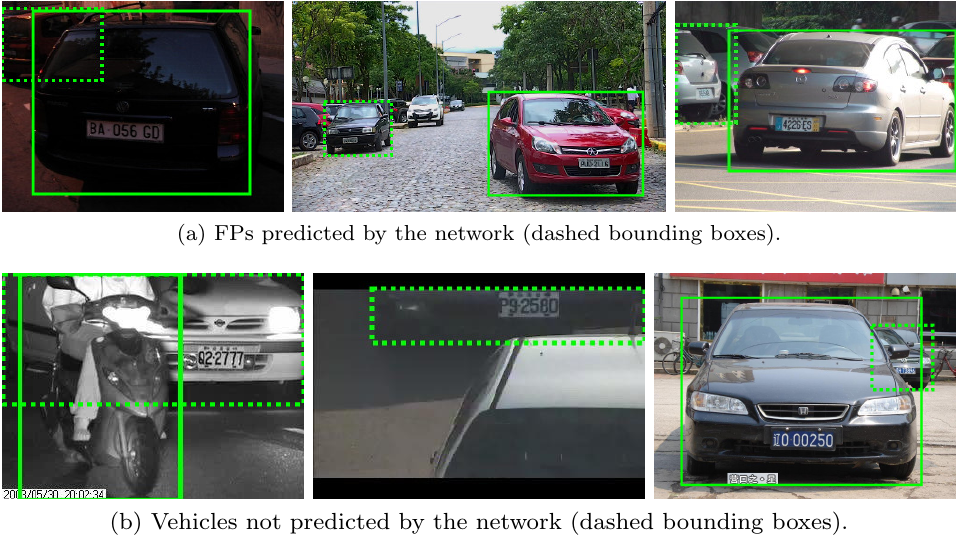}
    
    \vspace{-2mm}
    
    \caption{\gls*{fp} and \gls*{fn} predictions obtained in the vehicle detection stage. As can be seen in (a), the predicted \glspl*{fp} are mostly unlabelled vehicles in the background. In (b), one can see that the vehicles not predicted by the network (i.e.,~the \glspl*{fn}) are predominantly those occluded or in the~background.}
    \label{fig:results:vehicle_detection_fps_fns}
\end{figure}

The precision rates obtained by the network were only not higher due to unlabeled vehicles present in the background of the images, especially in the \aolp and \ssig datasets. 
Three examples are shown in Fig.~\ref{fig:results:vehicle_detection_fps_fns}a.
In Fig.~\ref{fig:results:vehicle_detection_fps_fns}b, we show some of the few cases where our network failed to detect one or more vehicles in the image. 
As can be seen, such cases are challenging since only a small part of each undetected vehicle is visible.

\subsection{License Plate Detection and Layout Classification}
\label{sec:results:lp_detection}

In Table~\ref{tab:results:lp_detection}, we report the results obtained by the modified Fast-YOLOv2 network in the \gls*{lp} detection and layout classification stage.
As we consider only one \gls*{lp} per vehicle image, the precision and recall rates are identical.
The average recall rate obtained in all datasets was $99.51$\% when disregarding the vehicles not detected in the previous stage and $99.45$\% when considering the entire test set.
This result is particularly impressive since we considered as incorrect the predictions in which the \gls*{lp}~layout was incorrectly classified with a high confidence value, even in cases where the \gls*{lp} position was predicted~correctly.

\begin{table}[!htb]
	\centering
	
	\captionsetup{position=top}
	
	\caption{Results attained in the \gls*{lp} detection and layout classification stage. The recall rates achieved in all datasets when disregarding the vehicles not detected in the previous stage are presented in~(a), while the recall rates obtained when considering the entire test set are listed in~(b).}
	\label{tab:results:lp_detection}
    \resizebox{0.9\columnwidth}{!}{
	\subfloat[][\label{tab:results:lp_detection_a}]{\begin{tabular}{@{}cc@{}}
		\toprule
		\textbf{Dataset} & \textbf{Recall (\%)} \\ \midrule
		\caltech & $99.13\pm1.19$ \\
		\englishlpd & $100.00\pm0.00$ \\
		\stills & $100.00\pm0.00$ \\
		\chinese & $100.00\pm0.00$ \\
		\aolp & $99.94\pm0.08$ \\
		\openalpreu & $98.52\pm0.51$ \\
		\ssig & $99.83\pm0.26$ \\
		\dataset & $98.67\pm0.25$ \\ \midrule
		\textbf{Average} & $\textbf{99.51}\boldsymbol{\pm}\textbf{0.29}$ \\ \bottomrule
	\end{tabular}} %
	\quad \vline \quad
	\subfloat[][\label{tab:results:lp_detection_b}]{\begin{tabular}{@{}cc@{}}
		\toprule
		\textbf{Dataset} & \textbf{Recall (\%)} \\ \midrule
		\caltech & $99.13\pm1.19$ \\
		\englishlpd & $100.00\pm0.00$ \\
		\stills & $100.00\pm0.00$ \\
		\chinese & $99.63\pm0.34$ \\
		\aolp & $99.85\pm0.10$ \\
		\openalpreu & $98.52\pm0.51$ \\
		\ssig & $99.80\pm0.24$ \\
		\dataset & $98.67\pm0.25$ \\ \midrule
		\textbf{Average} & $\textbf{99.45}\boldsymbol{\pm}\textbf{0.33}$ \\ \bottomrule
	\end{tabular}} \,}
\end{table}

\begin{figure}[!htb]
    \centering
    
    \includegraphics[width=0.99\columnwidth]{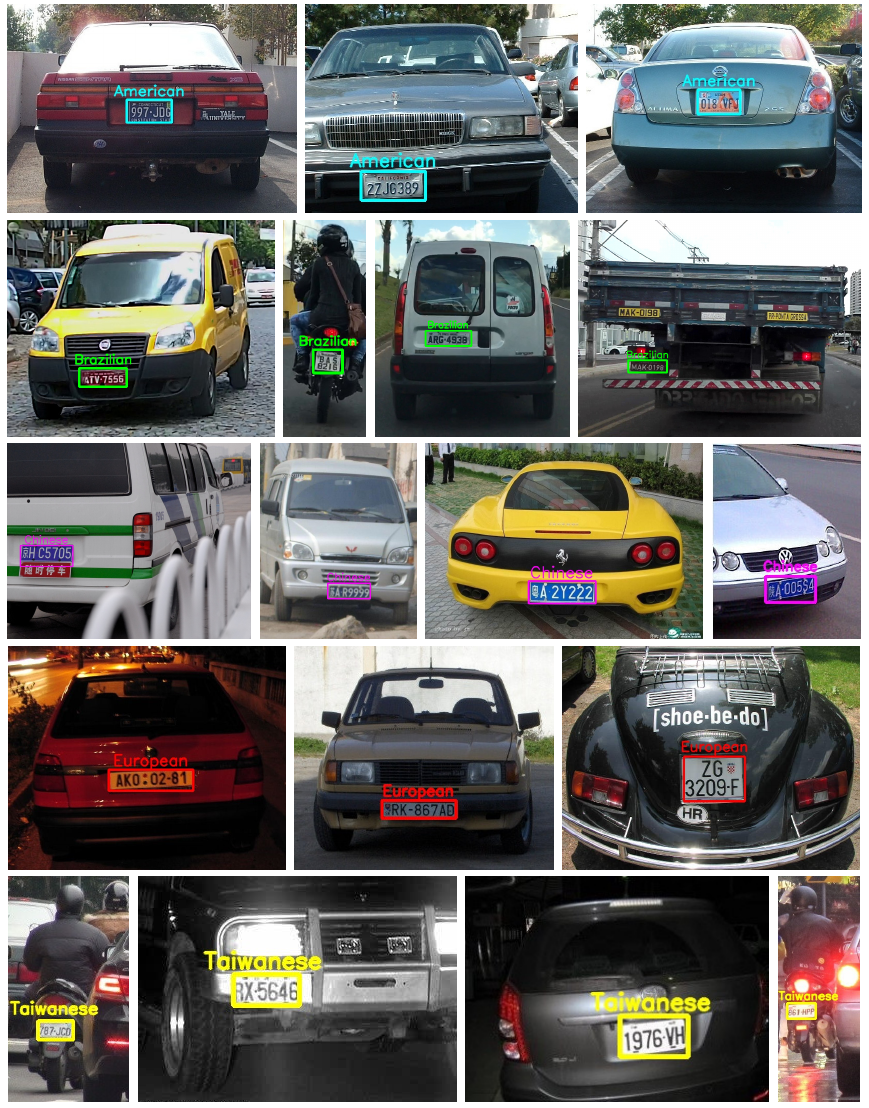}
    
    \vspace{-2mm}
    
    \caption[\glspl*{lp} correctly detected and classified by the proposed approach]{\glspl*{lp} correctly detected and classified by the proposed approach. Observe the robustness for this task regardless of vehicle type, lighting conditions, camera distance, and other factors.}
    \label{fig:results:lp_detection_tps}
    
\end{figure}

According to Fig.~\ref{fig:results:lp_detection_tps}, the proposed approach was able to successfully detect and classify \glspl*{lp} of various layouts, including those with few examples in the training set such as \glspl*{lp} issued in the U.S. states of Connecticut and Utah, or \glspl*{lp} of motorcycles \minor{registered in the Taiwan~region}.

\begin{figure}[!htb]
    \centering
    \captionsetup[subfloat]{captionskip=2pt,font={scriptsize}}
    
    \resizebox{\linewidth}{!}{
    \subfloat[][\colored{Examples of results obtained by detecting the \glspl*{lp} directly in the original image.} \label{fig:results:lp_detection_without_vehicle_detection:a}]{
    \includegraphics[height=11.5ex]{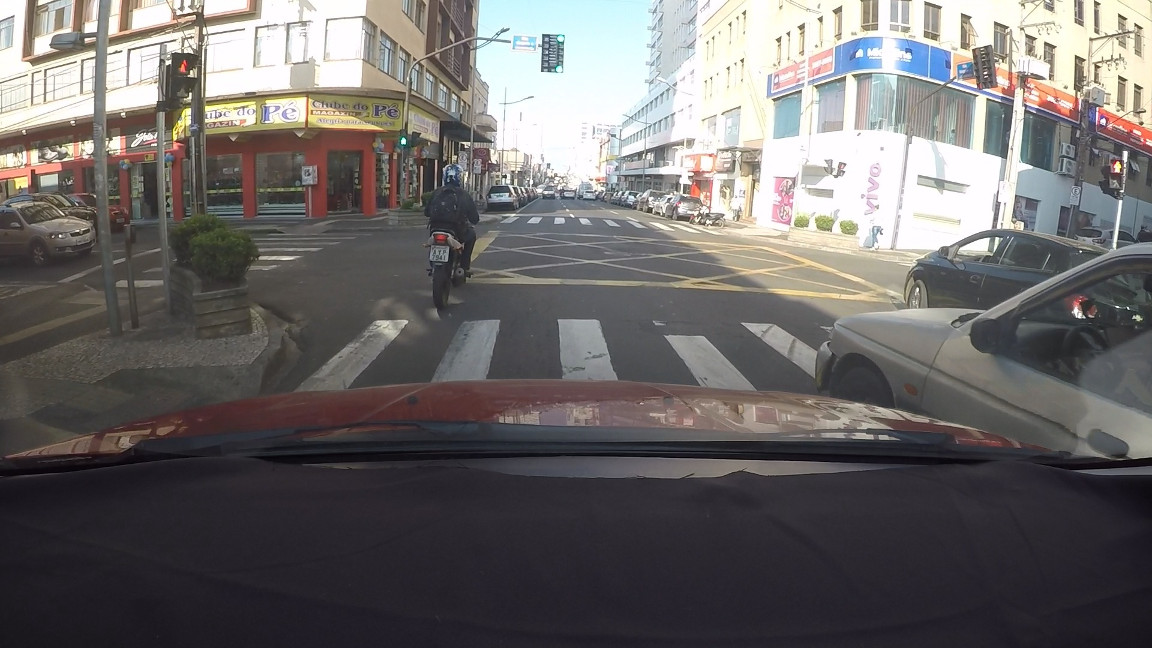}
    \includegraphics[height=11.5ex]{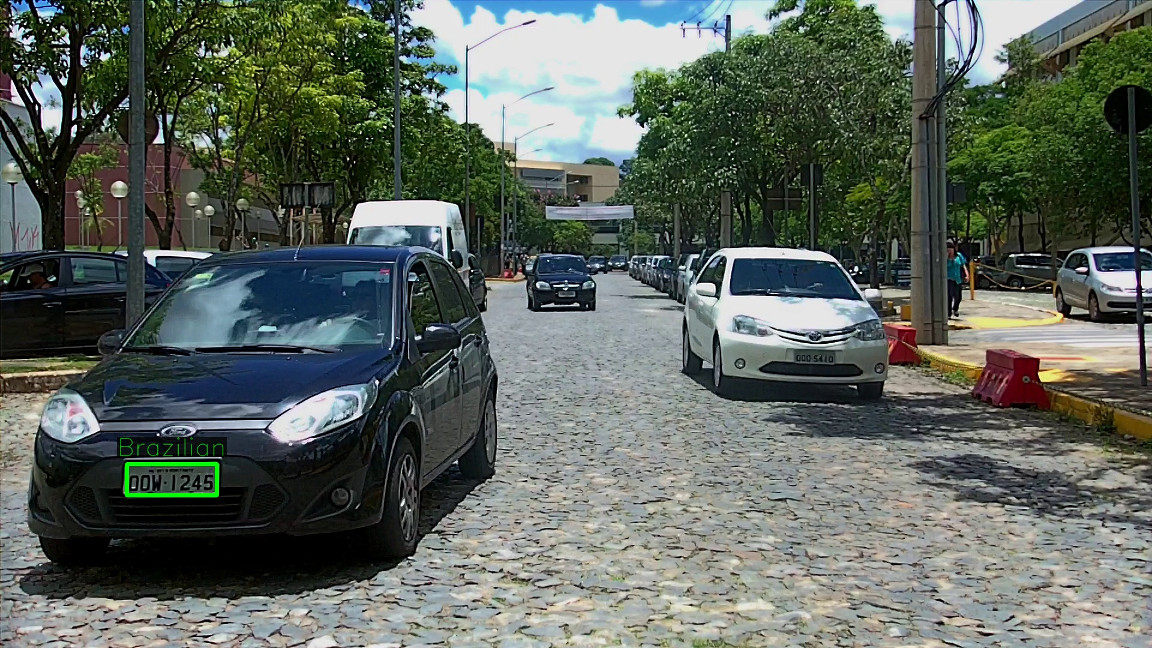}
    \includegraphics[height=11.5ex]{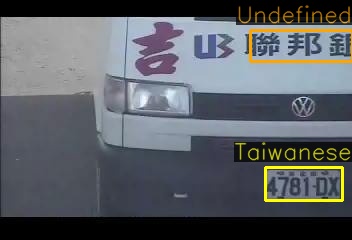}
    } \,
    }
    
    \vspace{1.5mm}
    
    \resizebox{\linewidth}{!}{
    \subfloat[][\colored{Examples of results obtained by detecting the \glspl*{lp} in the vehicle patches.}\label{fig:results:lp_detection_without_vehicle_detection:b}]{
    \includegraphics[height=11.5ex]{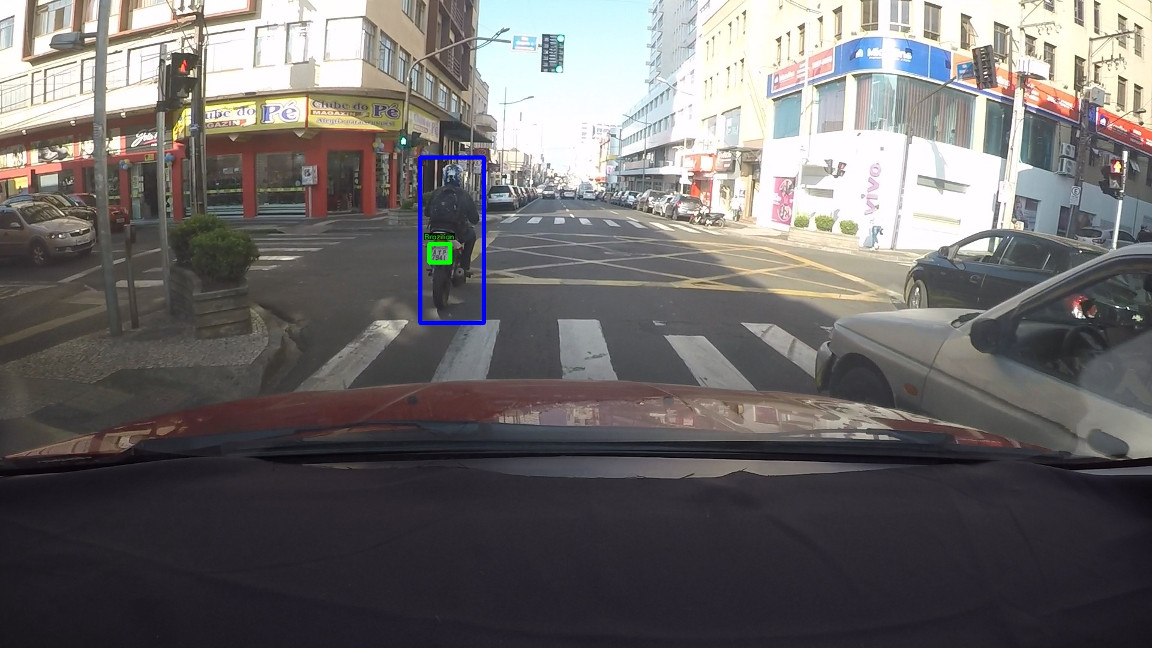}
    \includegraphics[height=11.5ex]{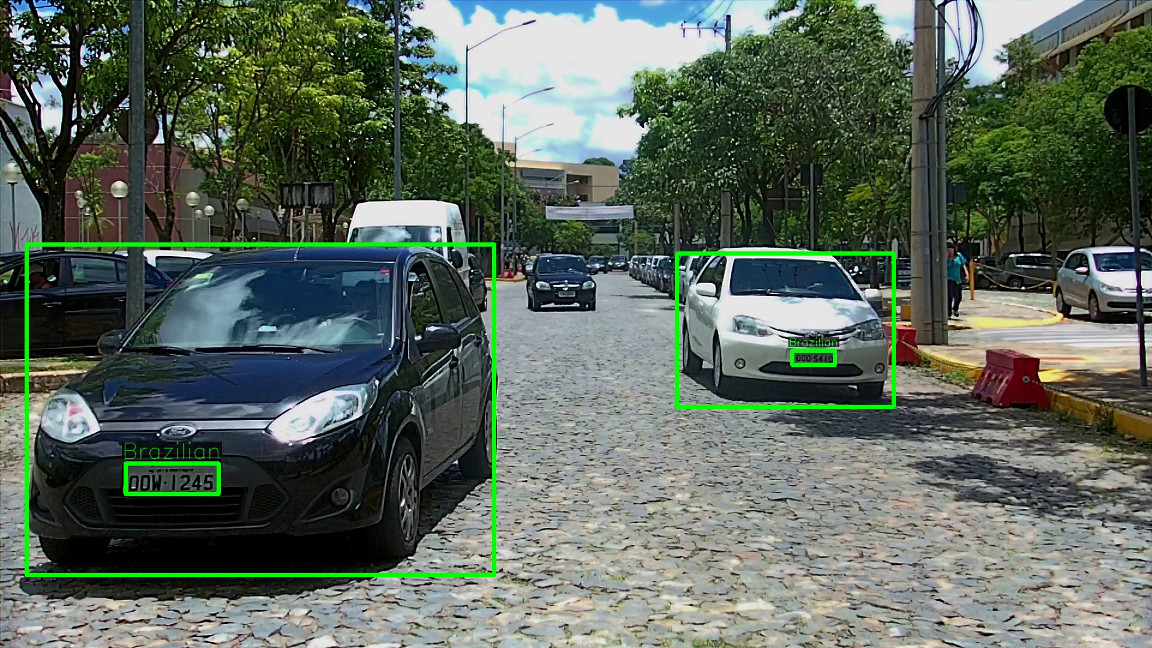}
    \includegraphics[height=11.5ex]{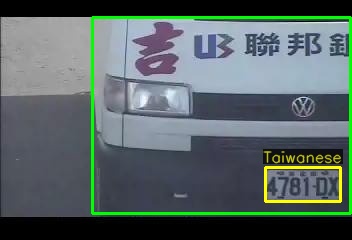}
    } \,
    } 
    
    \vspace{-1mm}
    
    \caption{\colored{Comparison of the results achieved by detecting/classifying the \glspl*{lp} directly in the original image~(a) and in the vehicle regions predicted in the vehicle detection stage~(b).}}
    \label{fig:results:lp_detection_without_vehicle_detection}
\end{figure}

\colored{It should be noted that (i)~the \glspl*{lp} may occupy a very small portion of the original image and that (ii)~textual blocks (e.g., phone numbers) on the vehicles or in the background can be confused with \glspl*{lp}.
Therefore, as can be seen in Fig.~\ref{fig:results:lp_detection_without_vehicle_detection}, the vehicle detection stage is \emph{\textbf{crucial}} for the effectiveness of our \gls*{alpr} system, as it helps to prevent both \glspl*{fp} and~\glspl*{fn}.}

\begin{figure}[!htb]
    \centering
    
    \includegraphics[width=0.99\columnwidth]{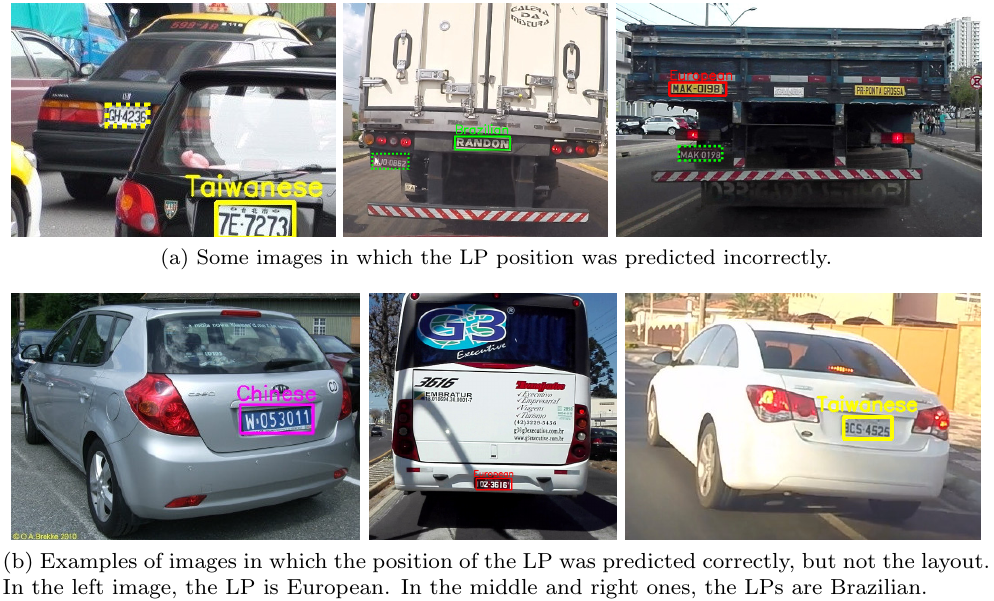}
    
    \vspace{-2mm}
    
    \caption[Some images in which our network failed either to detect the \gls*{lp} or to classify the \gls*{lp} layout]{Some images in which our network failed either to detect the \gls*{lp} or to classify the \gls*{lp} layout.}
    \label{fig:results:lp_detection_wrong}
\end{figure}

Some images where our network failed either to detect the \gls*{lp} or to classify the \gls*{lp} layout are shown in Fig.~\ref{fig:results:lp_detection_wrong}. 
As can be seen in Fig.~\ref{fig:results:lp_detection_wrong}a, our network failed to detect the \gls*{lp} in cases where there is a textual block very similar to an \gls*{lp} in the vehicle patch, or even when the \gls*{lp} of another vehicle appears within the patch (a single case in our experiments).
This is due to the fact that one vehicle can be almost totally occluded by another.
Regarding the errors in which the \gls*{lp} layout was misclassified, they occurred mainly in cases where the \gls*{lp} is considerably similar to \gls*{lp} of other layouts. 
For example, the left image in Fig.~\ref{fig:results:lp_detection_wrong}b shows a European \gls*{lp} (which has exactly the same colors and number of characters as standard Chinese \glspl*{lp}) incorrectly classified as Chinese.

It is important to note that it is still possible to correctly recognize the characters in some cases where our network has failed at this stage. For example, in the right image in Fig.~\ref{fig:results:lp_detection_wrong}a, the detected region contains exactly the same text as the ground truth (i.e., the~\gls*{lp}). Moreover, a Brazilian~\gls*{lp} classified as European (e.g., the middle image in Fig.~\ref{fig:results:lp_detection_wrong}b) can still be correctly recognized in the next stage since the only post-processing rule we apply to European~\glspl*{lp} is that they have between $5$ and $8$~characters.

As mentioned earlier, in this stage we disabled the color-related data augmentation of the Darknet framework. 
In this way, we eliminated more than half of the layout classification errors obtained when the model was trained using images with changed colors. 
We believe this is due to the fact that the network leverages color information (which may be distorted with some data augmentation approaches) for layout classification, as well as other characteristics such as the position of the characters and symbols on the~\gls*{lp}.

\subsection{License Plate Recognition (end-to-end)}
\label{sec:results:lp_recognition}

\begin{table*}[!htb]
\centering
\setlength{\tabcolsep}{9pt}
\caption{Recognition rates (\%) obtained by the proposed system, \colored{modified versions of our system}, previous works, and commercial systems in all datasets used in our experiments. To the best of our knowledge, in the literature, only algorithms for \gls*{lp} detection and character segmentation were evaluated in the \caltech, \stills and \chinese datasets. 
Therefore, our approaches are compared only with the commercial systems in these~datasets.
}
\vspace{1mm}

\label{tab:results:lp_recognition}
\resizebox{\textwidth}{!}{
\begin{tabular}{@{}ccccccccccc@{}}
\toprule
\diagbox[trim=l,trim=r,innerrightsep=-2.5pt,font=\footnotesize]{Dataset}{Approach} & \citep{panahi2017accurate} & \citep{zhuang2018towards} & \citep{silva2018license} & \colored{\citep{silva2020realtime}} & \citep{laroca2018robust} & Sighthound & OpenALPR & \colored{\begin{tabular}[c]{@{}c@{}}No Vehicle\\\phantom{i}Detection\footnotesize{$^\ast$}\end{tabular}} &  \colored{\begin{tabular}[c]{@{}c@{}}No Layout\\\phantom{i}Classification\footnotesize{$^\dagger$}\end{tabular}} & Proposed \\ \midrule
\caltech & $-$ & $-$ & $-$ & \colored{$-$} & $-$ & $95.7\pm2.7$ & $\textbf{99.1}\boldsymbol{\pm}\textbf{1.2}$ & \colored{$98.3\pm1.8$} & $96.1\pm1.8$ & $98.7\pm1.2$ \\
\englishlpd & $\textbf{97.0}$ & $-$ & $-$ & \colored{$-$} & $-$ & $92.5\pm3.7$ & $78.6\pm3.6$ & \colored{$95.3\pm1.6$} & $95.5\pm2.4$ & $95.7\pm2.3$ \\
\stills & $-$ & $-$ & $-$ & \colored{$-$} & $-$ & $\textbf{98.3}$ & $\textbf{98.3}$ & \colored{$98.0\pm0.7$} & $97.3\pm1.9$ & $98.0\pm1.4$ \\
\chinese & $-$ & $-$ & $-$ & \colored{$-$} & $-$ & $90.4\pm2.4$ & $92.6\pm1.9$ & \colored{$97.0\pm0.7$} & $95.4\pm1.1$ & $\textbf{97.5}\boldsymbol{\pm}\textbf{0.9}$ \\
\aolp & $-$ & $\textbf{99.8}$\footnotesize{$^{\ddagger}$} & $-$ & \colored{$-$} & $-$ & $87.1\pm0.8$ & $-$ & \colored{$98.8\pm0.3$} & $98.4\pm0.7$ & $99.2\pm0.4$ \\
\openalpreu & $-$ & $-$ & $93.5$ & \colored{$85.2$} & $-$ & $\colored{93.5}$ & $\colored{91.7}$ & \colored{$\textbf{97.8}\boldsymbol{\pm}\textbf{0.5}$} & $\colored{96.7\pm1.9}$ & $\colored{\textbf{97.8}\boldsymbol{\pm}\textbf{0.5}}$ \\
\ssig & $-$ & $-$ & $88.6$ & \colored{$89.2$} & $85.5$ & $82.8$ & $92.0$ & \colored{$96.5\pm0.9$} & $96.9\pm0.5$ & $\textbf{98.2}\boldsymbol{\pm}\textbf{0.5}$ \\
\dataset & $-$ & $-$ & $-$ & \colored{$-$} & $64.9$ & $62.3$ & $82.2$ & \colored{$59.6\pm0.9$} & $82.5\pm1.1$ & $\textbf{90.0}\boldsymbol{\pm}\textbf{0.7}$ \\ \midrule
Average & $-$ & $-$ & $-$ & \colored{$-$} & $-$ & $\colored{87.8\pm2.4}$ & $\colored{90.7\pm2.3}$ & \colored{$92.7\pm0.9$} & \colored{$94.8\pm1.4$} & $\colored{\textbf{96.9}\boldsymbol{\pm}\textbf{1.0}}$ \\ \bottomrule \\[-9pt]
\multicolumn{11}{l}{\footnotesize \colored{$^\ast$ A modified version of our approach in which the \glspl*{lp} are detected (and their layouts classified) directly in the original image (i.e., without vehicle detection).}} \\
\multicolumn{11}{l}{\footnotesize $^\dagger$ The proposed \gls*{alpr} system assuming that all \gls*{lp} layouts were classified as undefined (i.e., without layout classification and heuristic rules).} \\
\multicolumn{11}{l}{\footnotesize $^{\ddagger}$ The \gls*{lp} patches for the \gls*{lp} recognition stage were cropped directly from the ground truth in~\citep{zhuang2018towards}.}
\end{tabular}
}
\end{table*}

As in the vehicle detection stage, we first evaluated different confidence threshold values in the validation set in order to miss as few characters as possible, while avoiding high \gls*{fp}~rates.
We adopted a~$0.5$ confidence threshold for all \glspl*{lp} except European ones, where a higher threshold (i.e.,~$0.65$) was adopted since European \glspl*{lp} can have up to $8$ characters and several \glspl*{fp} were predicted on \glspl*{lp} with fewer characters when using a lower confidence threshold.

We considered the `$1$' and `I' characters as a single class in the assessments performed in the \ssig and \dataset datasets, as those characters are identical but occupy different positions on Brazilian~\glspl*{lp}.
The same procedure was done in~\citep{laroca2018robust,silva2018license}.

For each dataset, we compared the proposed \gls*{alpr} system with state-of-the-art methods that were evaluated using the same protocol as the one described in Section~\ref{sec:experiments:evaluation_protocol}. In addition, our results are compared with those obtained by Sighthound~\citep{masood2017sighthound} and OpenALPR~\citep{openalprapi}, which are two commercial systems often used as baselines in the \gls*{alpr} literature~\cite{spanhel2017holistic,laroca2018robust,goncalves2018realtime,silva2018license,silva2020realtime}.
According to the authors, both systems are robust for the detection and recognition of \glspl*{lp} of different layouts. 
It is important to emphasize that although the commercial systems were not tuned specifically for the datasets employed in our experiments, they are trained in much larger private datasets, which is a great advantage, especially in deep learning~approaches.

OpenALPR contains specialized solutions for \glspl*{lp} from different regions (e.g.,~\minor{mainland China}, Europe, among others) and the user must enter the correct region before using its~\acrshort*{api}, that is, it requires prior knowledge regarding the \gls*{lp}~layout. Sighthound, on the other hand, uses a single model/approach for \glspl*{lp} from different countries/regions, as well as the proposed system.

\colored{
The remainder of this section is divided into two parts.
First, in Section~\ref{sec:results:overall}, we conduct an overall evaluation of the proposed method across the eight datasets used in our experiments. 
The time required for our system to process an input image is also presented.
Afterward, in Section~\ref{sec:results:detailed}, we briefly present and discuss the results achieved by both the baselines and our \gls*{alpr} system on each dataset individually.
Such an analysis is very important to find out where the proposed system fails and the baselines do not, and vice versa.
}

\colored{
\subsection{Overall Evaluation}
\label{sec:results:overall}
}

The results obtained in all datasets by the proposed \gls*{alpr} system, previous works, and commercial systems are shown in Table~\ref{tab:results:lp_recognition}.
In the average of five runs, across all datasets, our end-to-end system correctly recognized \accavgmath of the \glspl*{lp}, outperforming Sighthound and OpenALPR by \outsighthound and \outopenalpr, respectively.
More specifically, the proposed system outperformed both previous works and commercial systems in the \chinese, \openalpreu, \ssig and \dataset datasets, and yielded competitive results to those attained by the~baselines in the other datasets. 

The proposed system attained results similar to those obtained by OpenALPR in the \caltech dataset ($98.7$\% against $99.1$\%, which represents a difference of less than one \gls*{lp} per run, on average, as there are only $46$ testing images), even though our system does not require prior knowledge. Regarding the \englishlpd dataset, our system performed better than the best baseline~\citep{panahi2017accurate} in~$2$ of the~$5$ runs \colored{(this evaluation highlights the importance of executing the proposed method five times and then averaging the results)}. Although we used the same number of images for testing, in~\citep{panahi2017accurate} the dataset was divided only once and the images used for testing were not specified. In the \stills dataset, both commercial systems reached a recognition rate of $98.3$\% while our system achieved $98$\% on average (with a standard deviation of $1.4$\%). 
Lastly, in the \aolp dataset, the proposed approach obtained similar results to those reported by~\cite{zhuang2018towards}, even though in their work the \gls*{lp} patches used as input in the \gls*{lp} recognition stage were cropped directly from the ground truth~(simplifying the problem, as explained in Section~\ref{sec:related_work}); in other words, they did not take into account vehicles or \glspl*{lp}  not detected in the earlier stages, nor background noise in the \gls*{lp} patches due to less accurate \gls*{lp}~detections.

\colored{To further highlight the importance of the vehicle detection stage, we included in Table~\ref{tab:results:lp_recognition} the results achieved by a modified version of our approach in which the \glspl*{lp} are detected (and their layouts classified) directly in the original image (i.e., without vehicle detection).
Although comparable results were achieved on datasets where the images were acquired on well-controlled scenarios, the modified version failed to detect/classify \glspl*{lp} in various images captured under less controlled conditions (as illustrated in Fig.~\ref{fig:results:lp_detection_without_vehicle_detection:b}), e.g. with vehicles far from the camera and shadows on the \glspl*{lp}, which explains the low recognition rate achieved by that approach in the challenging \dataset dataset -- where the images were taken from inside a vehicle driving through regular traffic in an urban environment, and most \glspl*{lp} occupy a very small region of the image~\cite{laroca2018robust}.}

Similarly, to evaluate the impact of classifying the \gls*{lp}~layout prior to \gls*{lp}~recognition (i.e., our main proposal), we also report in Table~\ref{tab:results:lp_recognition} the results obtained when assuming that all \gls*{lp} layouts were classified as undefined and that a generic approach (i.e., without heuristic rules) was employed in the \gls*{lp}~recognition stage.
The mean recognition rate was improved by~\improvelayoutclassification.
We consider this strategy (layout classification~+ heuristic rules) \emph{\textbf{essential}} for accomplishing outstanding results in datasets that contain \glspl*{lp} with fixed positions for letters and digits (e.g., Brazilian and Chinese \glspl*{lp}), as the recognition rates attained in the \chinese, \ssig and \dataset datasets were improved by~$3.6$\% on average.

The robustness of our \gls*{alpr} system is remarkable since it achieved recognition rates higher than $95$\% in all datasets except \dataset (\textit{where it outperformed the best baseline by $7.8$\%}).
The commercial systems, on the other hand, achieved similar results only in the \caltech and \stills datasets, which contain exclusively American~\glspl*{lp}, and performed poorly (i.e., recognition rates below $85$\%) in at least two datasets. 
This suggests that the commercial systems are not so well trained for \glspl*{lp} of other~layouts \minor{and highlights the importance of carrying out experiments on multiple datasets (with different characteristics) and not just on one or two, as is generally done in most works in the~literature.}

Although OpenALPR achieved better results than Sighthound (on average across all datasets), the latter system can be seen as more robust than the former since it does not require prior knowledge regarding the \gls*{lp} layout. 
In addition, OpenALPR does not support \minor{\glspl*{lp} from the Taiwan region}.
In this sense, we tried to employ OpenALPR solutions designed for \glspl*{lp} from other \minor{regions (including mainland China)} in the experiments performed in the \aolp~dataset; however, very low detection and recognition rates were obtained.

\begin{figure}[!htb]
	\centering
    
    \includegraphics[width=0.99\columnwidth]{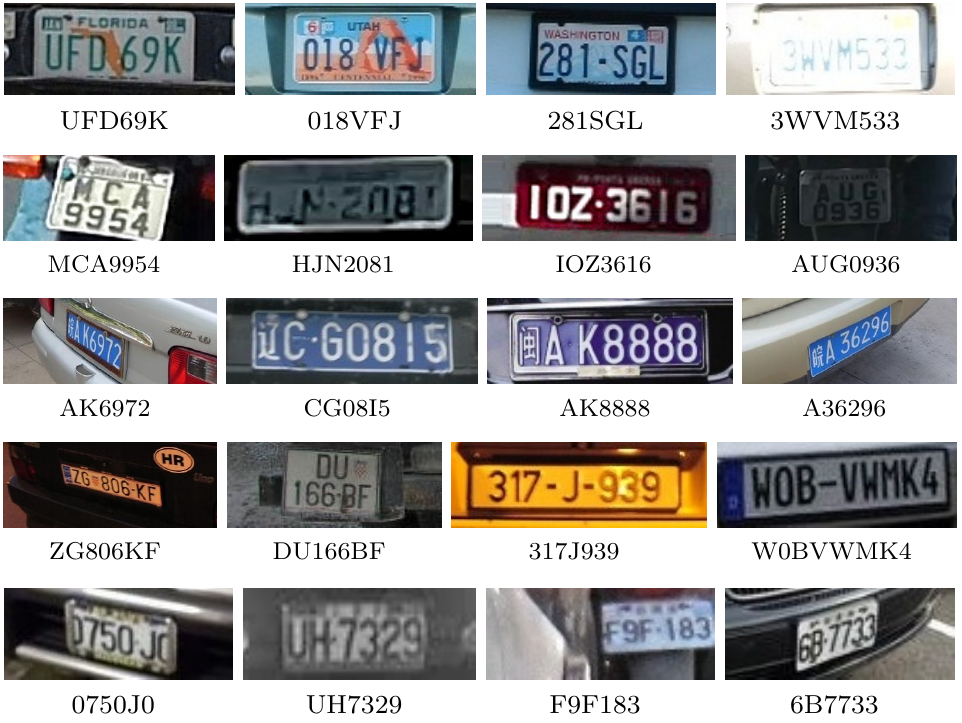}
    
    \vspace{-2mm}

	\caption{Examples of \glspl*{lp} that were correctly recognized by the proposed \gls*{alpr} system. From top to bottom: American, Brazilian, Chinese, European and Taiwanese~\glspl*{lp}.}
    \label{fig:results:lp_recognition_correct}  
\end{figure}

Fig.~\ref{fig:results:lp_recognition_correct} shows some examples of \glspl*{lp} that were correctly recognized by the proposed approach. As can be seen, our system can generalize well and correctly recognize \glspl*{lp} of different layouts, even when the images were captured under challenging conditions. It is noteworthy that, unlike~\cite{panahi2017accurate,laroca2018robust,zhuang2018towards}, the exact same networks were applied to all datasets; in other words, no specific training procedure was used to tune the networks for a given dataset or layout~class.
\colored{Instead, we use heuristic rules in cases where the LP layout is classified with a high confidence~value}.

\begin{figure}[!htb]
	\centering
    
    \includegraphics[width=0.99\columnwidth]{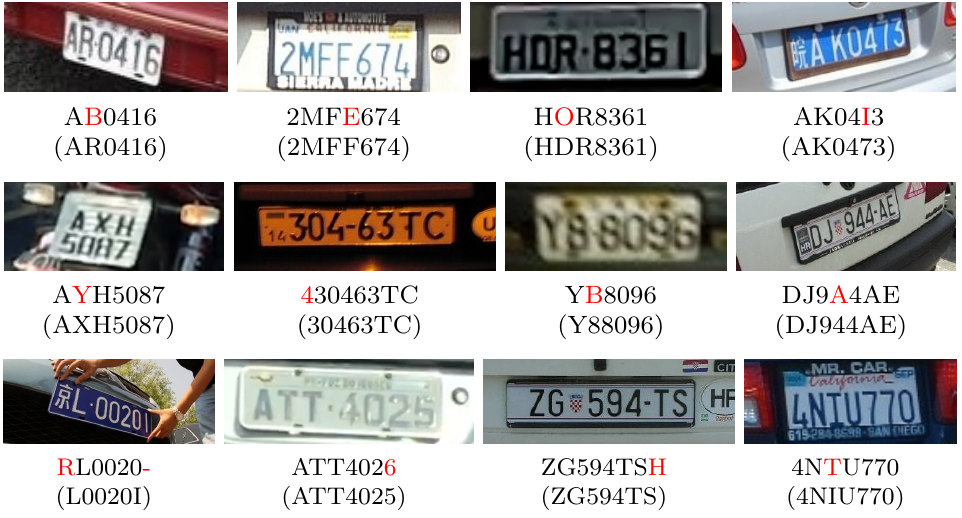}
    
    \vspace{-2mm}

	\caption{Examples of \glspl*{lp} that were incorrectly recognized by the proposed \gls*{alpr} system.
	The ground truth is shown in parentheses.}
    \label{fig:results:lp_recognition_wrong}  
\end{figure}

Some \glspl*{lp} in which our system failed to correctly detect/recognize all characters are shown in Fig.~\ref{fig:results:lp_recognition_wrong}. As one may see, the errors occurred mainly in challenging \gls*{lp} images, where even humans can make mistakes since, in some cases, one character might become very similar to another due to the inclination of the \gls*{lp}, the \gls*{lp} frame, shadows, blur, among other factors. 
Note that, in this work, we did not apply preprocessing techniques to the \gls*{lp} image in order not to increase the overall cost of the proposed system.

In Table~\ref{tab:times}, we report the time required for each network in our system to process an input.
As in~\citep{silva2017realtime,laroca2018robust,silva2020realtime}, the reported time is the average time spent processing all inputs in each stage, assuming that the network weights are already loaded and that there is a single vehicle in the scene.
\minor{Although a relatively deep model is explored for vehicle detection, our system is still able to process $73$~\gls*{fps} using a high-end~\acrshort*{gpu}.
In this sense, we believe that it can be employed for several real-world applications, such as parking and toll monitoring systems, even in cheaper setups (e.g., with a mid-end~GPU).}

\begin{table}[!htb]
	\centering
	\caption{The time required for each network in our system to process an input on an NVIDIA Titan Xp \acrshort*{gpu}.}
	\label{tab:times}
    
    \vspace{1mm}

    \resizebox{0.8\columnwidth}{!}{
    \begin{tabular}{@{}cccc@{}}
		\toprule
		\gls*{alpr} Stage & Adapted Model & Time (ms) & \gls*{fps} \\ \midrule
		Vehicle Detection & YOLOv2 &  $8.5382$ & $117$ \\[0.5ex]
		\multicolumn{1}{c}{\begin{tabular}[c]{@{}c@{}}\gls*{lp} Detection and\\[-0.3ex]Layout Classification\end{tabular}} & Fast-YOLOv2 & $3.0854$ & $324$ \\[1.8ex]
		\gls*{lp} Recognition & CR-NET & $1.9935$ & $502$ \\ \midrule
		\textbf{End-to-end} & $\textbf{-}$ & $\textbf{13.6171}$ & $\textbf{73}$ \\ \bottomrule
	\end{tabular}
	}
\end{table}

It should be noted that practically all images from the datasets used in our experiments contain only one labeled vehicle.
However, to perform a more realistic analysis of the execution time, we listed in Table~\ref{tab:times_more_lps} the time required for the proposed system to process images assuming that there is a certain number of vehicles in every image (note that vehicle detection is performed only once, regardless of the number of vehicles in the image).
According to the results, our system can process more than $30$ \gls*{fps} even when there are $4$ vehicles in the~scene.
This information is relevant since some \gls*{alpr} approaches, including the one proposed in our previous work~\citep{laroca2018robust}, can only run in real time if there is at most one vehicle in the~scene.

\begin{table}[!htb]
    \centering
    \caption{Execution times considering that there is a certain number of vehicles in every image.}
    \label{tab:times_more_lps}
    
    \vspace{1mm}
    
    \resizebox{0.45\columnwidth}{!}{ %
    \begin{tabular}{@{}ccc@{}}
    \toprule
    \# Vehicles & Time~(ms) & \gls*{fps} \\ \midrule
    $1$ & $13.6171$ & $73$ \\
    $2$ & $18.6960$ & $53$ \\
    $3$ & $23.7749$ & $42$ \\
    $4$ & $28.8538$ & $35$ \\[2pt] \cdashline{1-3} \\[-6pt]
    $5$ & $33.9327$ & $29$ \\ \bottomrule
    \end{tabular}
    }%
\end{table}

The proposed approach achieved an outstanding trade-off between accuracy and speed, unlike others recently proposed in the literature. For example, the methods proposed in~\citep{silva2017realtime,goncalves2018realtime} are capable of processing more images per second than our system but reached poor recognition rates (i.e., below $65$\%) in at least one dataset in which they were evaluated. On the other hand, impressive results were achieved on different scenarios in~\citep{li2018toward,li2018reading,silva2018license}.
However, the methods presented in these works are computationally expensive and cannot be applied in real time.
The Sighthound and OpenALPR commercial systems do not report the execution time.

\minor{We remark that real-time processing may be affected by many factors in practice.
For example, we measured our system's execution time when there was no other process consuming machine resources significantly. 
This is the standard procedure in the literature since it enables/facilitates the comparison of different approaches, despite the fact that it may not accurately represent some real-world applications, where other tasks must be performed simultaneously.
Some other factors that may affect real-time processing are the time it takes to transfer the image from the camera to the processing unit, hardware characteristics (e.g., CPU architecture, read/write speeds, and data transfer time between CPU and GPUs), and the versions of the frameworks and libraries used (e.g., OpenCV, Darknet and~CUDA).
}

\colored{
It is important to emphasize that, according to our experiments, the proposed \gls*{alpr} system is robust under different conditions while being efficient essentially due to the meticulous way in which we designed, optimized and combined its different parts, always seeking the best trade-off between accuracy and speed.
All strategies adopted are very important in some way for the robustness and/or efficiency of the proposed approach, and no specific part contributes more than the others in every scenario.
For example, as shown in Table~\ref{tab:results:lp_recognition} and Fig.~\ref{fig:results:lp_detection_without_vehicle_detection}, vehicle detection mainly helps to prevent false positives and false~negatives on complex scenarios, while layout classification (along with heuristic rules) mainly improves the recognition of \glspl*{lp} with a fixed number of characters and/or fixed positions for letters and~digits.
In the same way, both tasks and also \gls*{lp} recognition would not have been accomplished so successfully, or so efficiently, if not for careful modifications to the networks and exploration of data augmentation techniques (all details were given in Section~\ref{sec:proposed}).
}

\colored{
\subsection{Evaluation by Dataset}
\label{sec:results:detailed}
}

\colored{
In this section, we briefly discuss the results achieved by both the baselines and our \gls*{alpr} system on each dataset individually, striving to clearly identify what types of errors are generally made by each system.
For each dataset, we show some qualitative results obtained by the commercial systems and the proposed approach, since we know exactly which images/\glspl*{lp} these systems recognized correctly or not.
In the \openalpreu, \ssig and \dataset datasets, we also show some predictions obtained by the methods introduced in~\cite{silva2018license,laroca2018robust}, as their architectures and pre-trained weights were made publicly available by the respective authors.
Note that, as we are comparing different \gls*{alpr} systems, the \gls*{lp} images shown in this section were cropped directly from the ground~truth.
We focus on the recognition stage for visualization purposes and also because we consider this stage as the current bottleneck of \gls*{alpr} systems.
However, we pointed out cases where one or more systems did not return any predictions on multiple images from a given dataset, which may indicate that the \glspl*{lp} were not properly detected.\\
}

\begin{figure}[!htb]
    
    \centering
    \captionsetup[subfigure]{labelformat=empty,font={scriptsize}}
    
    \resizebox{\linewidth}{!}{
	\subfloat[][\centering \phantom{\,} \resizebox{\adj}{!}{\textbf{\phantom{1}\cite{masood2017sighthound}:}} \texttt{\textcolor{red}{L}3WAZ301}\hspace{\textwidth} \phantom{\,} \resizebox{\adj}{!}{\textbf{\phantom{1}\cite{openalprapi}:}} \texttt{\phantom{L}3WAZ301}\hspace{\textwidth} \phantom{\,} \resizebox{\adj}{!}{\textbf{Ours:}} \texttt{\phantom{L}3WAZ301}]{
		\includegraphics[width=0.23\linewidth]{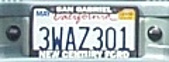}} \, %
    \subfloat[][\centering \phantom{\,} \resizebox{\adj}{!}{\textbf{\phantom{1}\cite{masood2017sighthound}:}} \texttt{\textcolor{red}{I}KCM356}\hspace{\textwidth} \phantom{\,} \resizebox{\adj}{!}{\textbf{\phantom{1}\cite{openalprapi}:}} \texttt{1KCM356}\hspace{\textwidth} \phantom{\,} \resizebox{\adj}{!}{\textbf{Ours:}} \texttt{1KCM356}]{
		\includegraphics[width=0.23\linewidth]{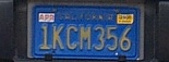}} \, %
	\subfloat[][\centering \phantom{\,} \resizebox{\adj}{!}{\textbf{\phantom{1}\cite{masood2017sighthound}:}} \texttt{2MFF674}\hspace{\textwidth} \phantom{\,} \resizebox{\adj}{!}{\textbf{\phantom{1}\cite{openalprapi}:}} \texttt{2MFF674}\hspace{\textwidth} \phantom{\,} \resizebox{\adj}{!}{\textbf{Ours:}} \texttt{2MF\textcolor{red}{E}674}]{
		\includegraphics[width=0.23\linewidth]{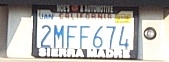}} \, %
	%
    \subfloat[][\centering \phantom{\,} \resizebox{\adj}{!}{\textbf{\phantom{1}\cite{masood2017sighthound}:}} \texttt{VZW818}\hspace{\textwidth} \phantom{\,} \resizebox{\adj}{!}{\textbf{\phantom{1}\cite{openalprapi}:}} \texttt{VZW818}\hspace{\textwidth} \phantom{\,} \resizebox{\adj}{!}{\textbf{Ours:}} \texttt{VZW818}]{
    		\includegraphics[width=0.23\linewidth]{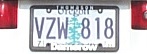}} \hspace{1.5mm}
	}
	
    \vspace{2.25mm}
	
	\resizebox{\linewidth}{!}{
	\subfloat[][\centering \phantom{\,} \resizebox{\adj}{!}{\textbf{\phantom{1}\cite{masood2017sighthound}:}} \texttt{997JDG}\hspace{\textwidth} \phantom{\,} \resizebox{\adj}{!}{\textbf{\phantom{1}\cite{openalprapi}:}} \texttt{997JDG}\hspace{\textwidth} \phantom{\,} \resizebox{\adj}{!}{\textbf{Ours:}} \texttt{997JDG}]{
		\includegraphics[width=0.23\linewidth]{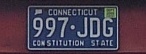}} \, %
    \subfloat[][\centering \phantom{\,} \resizebox{\adj}{!}{\textbf{\phantom{1}\cite{masood2017sighthound}:}} \texttt{4CY\textcolor{red}{2}275}\hspace{\textwidth} \phantom{\,} \resizebox{\adj}{!}{\textbf{\phantom{1}\cite{openalprapi}:}} \texttt{4CYE275}\hspace{\textwidth} \phantom{\,} \resizebox{\adj}{!}{\textbf{Ours:}} \texttt{4CYE275}]{
		\includegraphics[width=0.23\linewidth]{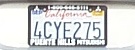}} \, %
	\subfloat[][\centering \phantom{\,} \resizebox{\adj}{!}{\textbf{\phantom{1}\cite{masood2017sighthound}:}} \texttt{VFY818}\hspace{\textwidth} \phantom{\,} \resizebox{\adj}{!}{\textbf{\phantom{1}\cite{openalprapi}:}} \texttt{VFY818}\hspace{\textwidth} \phantom{\,} \resizebox{\adj}{!}{\textbf{Ours:}} \texttt{VFY818}]{
		\includegraphics[width=0.23\linewidth]{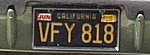}} \, %
    \subfloat[][\centering \phantom{aa} \resizebox{\adj}{!}{\textbf{\phantom{1}\cite{masood2017sighthound}:}} \texttt{\phantom{a}\textcolor{red}{F118}\phantom{a}}\phantom{i} \hspace{\textwidth} \phantom{aa} \resizebox{\adj}{!}{\textbf{\phantom{1}\cite{openalprapi}:}} \texttt{\phantom{aa}\textcolor{red}{n/a}\phantom{a}}\phantom{i} \hspace{\textwidth} \phantom{aa} \resizebox{\adj}{!}{\textbf{Ours:}} \texttt{\phantom{a}\textcolor{red}{IR69}\phantom{a}}\phantom{i}]{
		\includegraphics[width=0.23\linewidth]{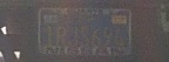}} \hspace{1.5mm}
	}
	
	\vspace{-0.5mm}

    \caption{\colored{Some qualitative results obtained on \caltech~\cite{caltech} by Sighthound~\cite{masood2017sighthound}, OpenALPR~\cite{openalprapi} and the proposed system.}}
    \label{fig:detailed-caltech}
\end{figure}

\colored{
\noindent
\textbf{\caltech~\cite{caltech}:} this is the dataset with fewer images for testing (only~$46$).
Hence, a single image recognized incorrectly reduces the accuracy of the system being evaluated by more than $2$\%.
By carefully analyzing the results, we found out that there is a challenging image in this dataset that neither the commercial systems nor the proposed system could correctly recognize.
Note that, in some executions, this image was not in the test subset, which explains the mean recognition rates above $98$\% attained by both our system and OpenALPR.
As illustrated in Fig.~\ref{fig:detailed-caltech}, while OpenALPR only made mistakes in that image, the proposed system failed in another image as well (where an~`F' looks like an~`E' due to the \gls*{lp}'s frame), and Sighthound failed in some other \glspl*{lp} due to very similar characters (e.g.,~`$1$'~and~`I') or false positives.\\
}

\begin{figure}[!htb]
    \centering
    \captionsetup[subfigure]{labelformat=empty,font={scriptsize}}
    
    \resizebox{\linewidth}{!}{
	\subfloat[][\centering \phantom{\,} \resizebox{\adj}{!}{\textbf{\phantom{1}\cite{masood2017sighthound}:}} \texttt{ZG200ID}\hspace{\textwidth} \phantom{\,} \resizebox{\adj}{!}{\textbf{\phantom{1}\cite{openalprapi}:}} \texttt{ZG200ID}\hspace{\textwidth} \phantom{\,} \resizebox{\adj}{!}{\textbf{Ours:}} \texttt{ZG200ID}]{
		\includegraphics[width=0.23\linewidth]{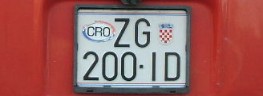}} \, %
    \subfloat[][\centering \phantom{\,} \resizebox{\adj}{!}{\textbf{\phantom{1}\cite{masood2017sighthound}:}} \texttt{ZG594TS\phantom{H}}\hspace{\textwidth} \phantom{\,} \resizebox{\adj}{!}{\textbf{\phantom{1}\cite{openalprapi}:}} \texttt{ZG594TS\phantom{H}}\hspace{\textwidth} \phantom{\,} \resizebox{\adj}{!}{\textbf{Ours:}} \texttt{ZG594TS\textcolor{red}{H}}]{
		\includegraphics[width=0.23\linewidth]{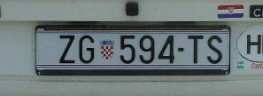}} \, %
	\subfloat[][\centering \phantom{\,} \resizebox{\adj}{!}{\textbf{\phantom{1}\cite{masood2017sighthound}:}} \texttt{ZG511S\textcolor{red}{-}}\hspace{\textwidth} \phantom{\,} \resizebox{\adj}{!}{\textbf{\phantom{1}\cite{openalprapi}:}} \texttt{ZG511\textcolor{red}{9-}}\hspace{\textwidth} \phantom{\,} \resizebox{\adj}{!}{\textbf{Ours:}} \texttt{ZG511SF}]{
		\includegraphics[width=0.23\linewidth]{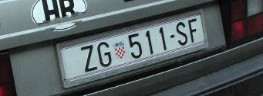}} \, %
	%
    \subfloat[][\centering \phantom{a} \resizebox{\adj}{!}{\textbf{\phantom{1}\cite{masood2017sighthound}:}} \texttt{AHV\textcolor{red}{8}9002}\phantom{i} \hspace{\textwidth} \phantom{a} \resizebox{\adj}{!}{\textbf{\phantom{1}\cite{openalprapi}:}} \texttt{\phantom{aa}\textcolor{red}{n/a}\phantom{aaa}}\phantom{i} \hspace{\textwidth} \phantom{a} \resizebox{\adj}{!}{\textbf{Ours:}} \texttt{AHV\textcolor{red}{8}9002}\phantom{i}]{
		\includegraphics[width=0.23\linewidth]{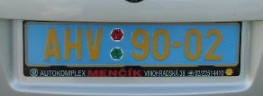}} \hspace{1.5mm}
	}
	
    \vspace{2.25mm}
	
	\resizebox{\linewidth}{!}{
	\subfloat[][\centering \phantom{\,} \resizebox{\adj}{!}{\textbf{\phantom{1}\cite{masood2017sighthound}:}} \texttt{VU279A\textcolor{red}{-}}\hspace{\textwidth} \phantom{\,} \resizebox{\adj}{!}{\textbf{\phantom{1}\cite{openalprapi}:}} \texttt{VU279AE}\hspace{\textwidth} \phantom{\,} \resizebox{\adj}{!}{\textbf{Ours:}} \texttt{VU279AE}]{
		\includegraphics[width=0.23\linewidth]{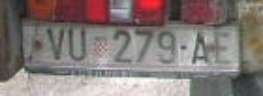}} \, %
	\subfloat[][\centering \phantom{\,} \resizebox{\adj}{!}{\textbf{\phantom{1}\cite{masood2017sighthound}:}} \texttt{HGAS1802}\hspace{\textwidth} \phantom{\,} \resizebox{\adj}{!}{\textbf{\phantom{1}\cite{openalprapi}:}} \texttt{HGAS180\textcolor{red}{-}}\hspace{\textwidth} \phantom{\,} \resizebox{\adj}{!}{\textbf{Ours:}} \texttt{HGAS1802}]{
		\includegraphics[width=0.23\linewidth]{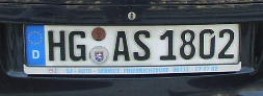}} \, %
	\subfloat[][\centering  \phantom{\,} \resizebox{\adj}{!}{\textbf{\phantom{1}\cite{masood2017sighthound}:}}~\texttt{RI393BD}\hspace{\textwidth} \phantom{\,} \resizebox{\adj}{!}{\textbf{\phantom{1}\cite{openalprapi}:}}~\texttt{RI393BD}\hspace{\textwidth} \phantom{\,} \resizebox{\adj}{!}{\textbf{Ours:}}~\texttt{RI393BD}]{
		\includegraphics[width=0.23\linewidth]{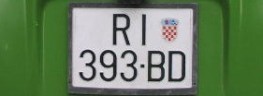}} \, %
    \subfloat[][\centering  \phantom{\,} \resizebox{\adj}{!}{\textbf{\phantom{1}\cite{masood2017sighthound}:}}~\texttt{VZ876C\textcolor{red}{T}}\hspace{\textwidth} \phantom{\,} \resizebox{\adj}{!}{\textbf{\phantom{1}\cite{openalprapi}:}}~\texttt{VZ876C\textcolor{red}{T}}\hspace{\textwidth} \phantom{\,} \resizebox{\adj}{!}{\textbf{Ours:}}~\texttt{V\textcolor{red}{2}876C\textcolor{red}{1}}]{
		\includegraphics[width=0.23\linewidth]{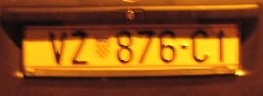}} \hspace{1.5mm}
	}
	
	\vspace{-0.5mm}

    \caption{\colored{Some qualitative results obtained on \englishlpd~\cite{englishlpd} by Sighthound~\cite{masood2017sighthound}, OpenALPR~\cite{openalprapi} and the proposed system.}}
    \label{fig:detailed-english}
\end{figure}

\colored{
\noindent \textbf{\englishlpd~\cite{englishlpd}:} this dataset has several \gls*{lp} layouts and different types of vehicles such as cars, buses and trucks.
Panahi \& Gholampour~\cite{panahi2017accurate} reported a recognition rate of $97.0$\% in this dataset, however, their method was executed only once and the images used for testing were not specified.
As can be seen in Table~\ref{tab:detailed-english}, using the same number of test images, our method achieved recognition rates above $97$\% in two out of five executions (Sighthound also surpassed $97$\% in one run).
In this sense, we consider that our system is as robust as the one presented in~\cite{panahi2017accurate}.
According to Fig.~\ref{fig:detailed-english}, neither the commercial systems nor the proposed system had difficulty in recognizing \glspl*{lp} with two rows of characters in this dataset.
Instead, as there are many different \gls*{lp} layouts in Europe and thus the number of characters on each \gls*{lp} is not fixed, most errors refer to a character being lost (i.e., false negatives) or, conversely, a non-existent character being predicted (i.e., false positives).
The low recognition rates achieved by OpenALPR are due to the fact that it did not return any predictions in some cases (as if there were no vehicles/\glspl*{lp} in the image).
In this sense, we conjecture that OpenALPR only returns predictions obtained with a high confidence value and that it is not as well trained for European \glspl*{lp} as it is for American/Brazilian~ones.
}

\begin{table}[!htb]
\centering
\setlength{\tabcolsep}{8pt}

\caption{\colored{Recognition rates (\%) achieved by Panahi \& Gholampour~\cite{panahi2017accurate}, Sighthound~\cite{masood2017sighthound}, OpenALPR~\cite{openalprapi}, and our system on \englishlpd~\cite{englishlpd}.}}
\label{tab:detailed-english}
\vspace{1mm}
\colored{
\resizebox{0.65\linewidth}{!}{%
\begin{tabular}{@{}ccccc@{}}
\toprule
Run & \cite{panahi2017accurate} & \cite{masood2017sighthound} & \cite{openalprapi} & Proposed \\ \midrule
\# $1$   & $-$     & $\textbf{98.0}$       & $82.4$     & $96.1$       \\
\# $2$   & $-$     & $94.1$       & $79.4$     &  $\textbf{97.1}$      \\
\# $3$   & $-$     & $91.2$          & $76.5$         &   $\textbf{98.0}$     \\
\# $4$   & $-$     & $91.2$       & $73.5$     & $\textbf{95.1}$       \\
\# $5$   & $-$          &  $88.2$          &   $81.4$       &   $\textbf{92.2}$     \\ \midrule
Average &  $\textbf{97.0}$        &  $92.5$          &  $78.6$        &  $95.7$       \\ \bottomrule
\end{tabular}
}%
}
\end{table}

\begin{figure}[!htb]
    \centering
    \captionsetup[subfigure]{labelformat=empty,font={scriptsize}}
    
    \resizebox{\linewidth}{!}{
	\subfloat[][\centering \phantom{\,} \resizebox{\adj}{!}{\textbf{\phantom{1}\cite{masood2017sighthound}:}}  \texttt{C\textcolor{red}{N}C3951}\hspace{\textwidth} \phantom{\,} \resizebox{\adj}{!}{\textbf{\phantom{1}\cite{openalprapi}:}} \texttt{CKC3951}\hspace{\textwidth}  \phantom{\,} \resizebox{\adj}{!}{\textbf{Ours:}} \texttt{CKC3951}]{
		\includegraphics[width=0.23\linewidth]{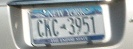}} \, %
    \subfloat[][\centering \phantom{\,} \resizebox{\adj}{!}{\textbf{\phantom{1}\cite{masood2017sighthound}:}} \texttt{RNM25X}\hspace{\textwidth} \phantom{\,} \resizebox{\adj}{!}{\textbf{\phantom{1}\cite{openalprapi}:}} \texttt{RNM25X}\hspace{\textwidth} \phantom{\,} \resizebox{\adj}{!}{\textbf{Ours:}} \texttt{RNM25X}]{
		\includegraphics[width=0.23\linewidth]{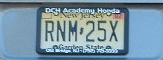}} \, %
	\subfloat[][\centering \phantom{\,} \resizebox{\adj}{!}{\textbf{\phantom{1}\cite{masood2017sighthound}:}} \texttt{1TM115\phantom{a}}\hspace{\textwidth} \phantom{\,} \resizebox{\adj}{!}{\textbf{\phantom{1}\cite{openalprapi}:}} \texttt{1TM115\phantom{a}}\hspace{\textwidth} \phantom{\,} \resizebox{\adj}{!}{\textbf{Ours:}} \texttt{1TM115\textcolor{red}{I}}]{
		\includegraphics[width=0.23\linewidth]{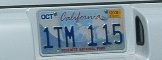}} \, %
    \subfloat[][\centering \phantom{\,} \resizebox{\adj}{!}{\textbf{\phantom{1}\cite{masood2017sighthound}:}} \texttt{5CGP522}\hspace{\textwidth} \phantom{\,} \resizebox{\adj}{!}{\textbf{\phantom{1}\cite{openalprapi}:}} \texttt{5CGP522}\hspace{\textwidth} \phantom{\,} \resizebox{\adj}{!}{\textbf{Ours:}} \texttt{5CGP522}]{
		\includegraphics[width=0.23\linewidth]{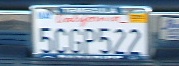}} \hspace{1.5mm}
	}
	
    \vspace{2.25mm}
	
	\resizebox{\linewidth}{!}{
	\subfloat[][\centering \phantom{\,} \resizebox{\adj}{!}{\textbf{\phantom{1}\cite{masood2017sighthound}:}}  \texttt{3J66282}\hspace{\textwidth} \phantom{\,} \resizebox{\adj}{!}{\textbf{\phantom{1}\cite{openalprapi}:}} \texttt{3J66282}\hspace{\textwidth}  \phantom{\,} \resizebox{\adj}{!}{\textbf{Ours:}} \texttt{3J66282}]{
		\includegraphics[width=0.23\linewidth]{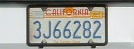}} \, %
    \subfloat[][\centering \phantom{\,} \resizebox{\adj}{!}{\textbf{\phantom{1}\cite{masood2017sighthound}:}} \texttt{4NFU116}\hspace{\textwidth} \phantom{\,} \resizebox{\adj}{!}{\textbf{\phantom{1}\cite{openalprapi}:}} \texttt{4NF\textcolor{red}{-}116}\hspace{\textwidth} \phantom{\,} \resizebox{\adj}{!}{\textbf{Ours:}} \texttt{4NFU116}]{
		\includegraphics[width=0.23\linewidth]{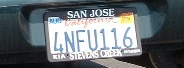}} \, %
	\subfloat[][\centering \phantom{\,} \resizebox{\adj}{!}{\textbf{\phantom{1}\cite{masood2017sighthound}:}} \texttt{AHA6497}\hspace{\textwidth} \phantom{\,} \resizebox{\adj}{!}{\textbf{\phantom{1}\cite{openalprapi}:}} \texttt{AHA6497}\hspace{\textwidth} \phantom{\,} \resizebox{\adj}{!}{\textbf{Ours:}} \texttt{AHA6497}]{
		\includegraphics[width=0.23\linewidth]{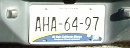}} \, %
    \subfloat[][\centering \phantom{\,} \resizebox{\adj}{!}{\textbf{\phantom{1}\cite{masood2017sighthound}:}} \texttt{4NIU770}\hspace{\textwidth} \phantom{\,} \resizebox{\adj}{!}{\textbf{\phantom{1}\cite{openalprapi}:}} \texttt{4NIU770}\hspace{\textwidth} \phantom{\,} \resizebox{\adj}{!}{\textbf{Ours:}} \texttt{4N\textcolor{red}{T}U770}]{
		\includegraphics[width=0.23\linewidth]{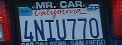}} \hspace{1.5mm}
	}
	
	\vspace{-0.5mm}

    \caption{\colored{Some qualitative results obtained on \stills~\cite{ucsd} by Sighthound~\cite{masood2017sighthound}, OpenALPR~\cite{openalprapi} and the proposed system.}}
    \label{fig:detailed-stills}
\end{figure}

\colored{
\noindent \textbf{\stills~\cite{ucsd}:} as \caltech, the \stills dataset also has few test images (only $60$).
Despite containing \glspl*{lp} from distinct U.S. states (i.e., different \gls*{lp} layouts) and under several lighting conditions, all \gls*{alpr} systems evaluated by us achieved excellent results in this dataset.
More specifically, both Sighthound and OpenALPR failed in just one image (interestingly, not in the same one).
This is another indication that these commercial systems are very well trained for American \glspl*{lp}.
Also very robustly, our system failed in just two images \underline{over $5$ runs}, remarkably recognizing all $60$ images correctly in one of them.
All images in which at least one system failed, as well as other representative ones, are shown in Fig.~\ref{fig:detailed-stills}.\\
}

\begin{figure}[!htb]
    \centering
    \captionsetup[subfigure]{labelformat=empty,font={scriptsize}}
    
    \resizebox{\linewidth}{!}{
	\subfloat[][\centering \phantom{\,} \resizebox{\adj}{!}{\textbf{\phantom{1}\cite{masood2017sighthound}:}} \texttt{ALA8\textcolor{red}{-{}-}}\hspace{\textwidth} \phantom{\,} \resizebox{\adj}{!}{\textbf{\phantom{1}\cite{openalprapi}:}} \texttt{A\textcolor{red}{I}A82I}\hspace{\textwidth} \phantom{\,} \resizebox{\adj}{!}{\textbf{Ours:}} \texttt{ALA82I}]{
		\includegraphics[width=0.23\linewidth]{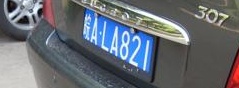}} \, %
    \subfloat[][\centering \phantom{\,} \resizebox{\adj}{!}{\textbf{\phantom{1}\cite{masood2017sighthound}:}} \texttt{ADT444}\hspace{\textwidth} \phantom{\,} \resizebox{\adj}{!}{\textbf{\phantom{1}\cite{openalprapi}:}} \texttt{AD\textcolor{red}{I}444}\hspace{\textwidth} \phantom{\,} \resizebox{\adj}{!}{\textbf{Ours:}} \texttt{AD\textcolor{red}{I}444}]{
		\includegraphics[width=0.23\linewidth]{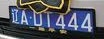}} \, %
	\subfloat[][\centering \phantom{\,} \resizebox{\adj}{!}{\textbf{\phantom{1}\cite{masood2017sighthound}:}} \texttt{\textcolor{red}{B}ABII57}\hspace{\textwidth} \phantom{\,} \resizebox{\adj}{!}{\textbf{\phantom{1}\cite{openalprapi}:}} \texttt{\phantom{a}ABII57}\hspace{\textwidth} \phantom{\,} \resizebox{\adj}{!}{\textbf{Ours:}} \texttt{\phantom{a}ABII57}]{
		\includegraphics[width=0.23\linewidth]{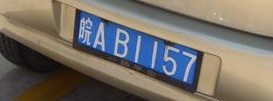}} \, %
    \subfloat[][\centering \phantom{\,} \resizebox{\adj}{!}{\textbf{\phantom{1}\cite{masood2017sighthound}:}} \texttt{AK0473}\hspace{\textwidth} \phantom{\,} \resizebox{\adj}{!}{\textbf{\phantom{1}\cite{openalprapi}:}} \texttt{AK0473}\hspace{\textwidth} \phantom{\,} \resizebox{\adj}{!}{\textbf{Ours:}} \texttt{AK04\textcolor{red}{I}3}]{
		\includegraphics[width=0.23\linewidth]{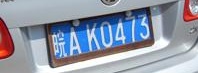}} \hspace{1.5mm}
	}
	
    \vspace{2.25mm}
	
	\resizebox{\linewidth}{!}{
	\subfloat[][\centering \phantom{\,} \resizebox{\adj}{!}{\textbf{\phantom{1}\cite{masood2017sighthound}:}} \texttt{C44444}\hspace{\textwidth} \phantom{\,} \resizebox{\adj}{!}{\textbf{\phantom{1}\cite{openalprapi}:}} \texttt{\textcolor{red}{G}44444}\hspace{\textwidth} \phantom{\,} \resizebox{\adj}{!}{\textbf{Ours:}} \texttt{C44444}]{
		\includegraphics[width=0.23\linewidth]{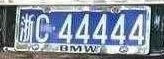}} \, %
	\subfloat[][\centering \phantom{\,} \resizebox{\adj}{!}{\textbf{\phantom{1}\cite{masood2017sighthound}:}} \texttt{AEI4\textcolor{red}{L}I}\hspace{\textwidth} \phantom{\,} \resizebox{\adj}{!}{\textbf{\phantom{1}\cite{openalprapi}:}} \texttt{AEI4II}\hspace{\textwidth} \phantom{\,} \resizebox{\adj}{!}{\textbf{Ours:}} \texttt{AEI4II}]{
		\includegraphics[width=0.23\linewidth]{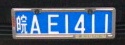}} \, %
	%
	\subfloat[][\centering \phantom{\,} \resizebox{\adj}{!}{\textbf{\phantom{1}\cite{masood2017sighthound}:}} \texttt{\textcolor{red}{I}A6ITII}\hspace{\textwidth} \phantom{\,} \resizebox{\adj}{!}{\textbf{\phantom{1}\cite{openalprapi}:}} \texttt{\phantom{a}A6ITI\textcolor{red}{7}}\hspace{\textwidth} \phantom{\,} \resizebox{\adj}{!}{\textbf{Ours:}} \texttt{\phantom{a}A6ITII}]{
		\includegraphics[width=0.23\linewidth]{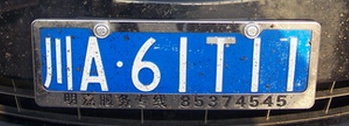}} \, %
    \subfloat[][\centering  \phantom{\,} \resizebox{\adj}{!}{\textbf{\phantom{1}\cite{masood2017sighthound}:}}~\texttt{\textcolor{red}{R}L0020I}\hspace{\textwidth} \phantom{\,} \resizebox{\adj}{!}{\textbf{\phantom{1}\cite{openalprapi}:}}~\texttt{\phantom{a}L0020\textcolor{red}{N}}\hspace{\textwidth} \phantom{\,} \resizebox{\adj}{!}{\textbf{Ours:}}~\texttt{\textcolor{red}{R}L0020\textcolor{red}{-}}]{
		\includegraphics[width=0.23\linewidth]{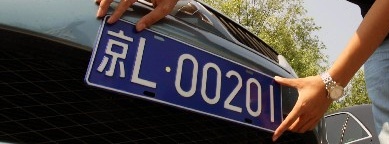}} \hspace{1.5mm}
	}
	
	\vspace{-0.5mm}

    \caption{\colored{Some qualitative results obtained on \chinese~\cite{zhou2012principal} by Sighthound~\cite{masood2017sighthound}, OpenALPR~\cite{openalprapi} and the proposed system.}}
    \label{fig:detailed-chinese}
\end{figure}

\colored{
\noindent \textbf{\chinese~\cite{zhou2012principal}:} this dataset contains both images captured by the authors and downloaded from the Internet. We used $159$ images for testing in each run.
An important feature of \chinese is that it has several images in which the \glspl*{lp} are tilted or inclined, as shown in Fig.~\ref{fig:detailed-chinese}.
In fact, most of the prediction errors obtained by commercial systems were in such images.
Our system, on the other hand, handled tilted/inclined \glspl*{lp} well and mostly failed in cases where one character become
very similar to another due to the \gls*{lp}
frame, shadows, blur, etc. 
It should be noted that Sighthound~($90.4$\%) misclassified the Chinese character (see Section~\ref{sec:proposed:lp_recognition} for details) as an English letter on some occasions.
This kind of recognition error was rarely made by the proposed system~($97.5$\%) and OpenALPR~($92.6$\%).
}

\begin{figure}[!htb]
    \centering
    \captionsetup[subfigure]{labelformat=empty,font={scriptsize}}
    
    \resizebox{\linewidth}{!}{
	\subfloat[][\centering \phantom{\,} \resizebox{\adj}{!}{\textbf{\phantom{1}\cite{masood2017sighthound}:}} \texttt{C\textcolor{red}{8}8117}\hspace{\textwidth} \phantom{\,} \resizebox{\adj}{!}{\textbf{Ours:}} \texttt{C38117}]{
		\includegraphics[width=0.23\linewidth]{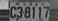}} \, %
    \subfloat[][\centering \phantom{\,} \resizebox{\adj}{!}{\textbf{\phantom{1}\cite{masood2017sighthound}:}} \texttt{RE9302}\hspace{\textwidth} \phantom{\,}  \resizebox{\adj}{!}{\textbf{Ours:}} \texttt{R\textcolor{red}{F}9302}]{
		\includegraphics[width=0.23\linewidth]{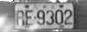}} \, %
	\subfloat[][\centering \phantom{\,} \resizebox{\adj}{!}{\textbf{\phantom{1}\cite{masood2017sighthound}:}} \texttt{8695LS}\hspace{\textwidth} \phantom{\,}  \resizebox{\adj}{!}{\textbf{Ours:}} \texttt{8695LS}]{
		\includegraphics[width=0.23\linewidth]{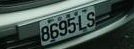}} \, %
    \subfloat[][\centering \phantom{\,} \resizebox{\adj}{!}{\textbf{\phantom{1}\cite{masood2017sighthound}:}} \texttt{Y\textcolor{red}{B}8096}\hspace{\textwidth} \phantom{\,}  \resizebox{\adj}{!}{\textbf{Ours:}} \texttt{Y\textcolor{red}{B}8096}]{
		\includegraphics[width=0.23\linewidth]{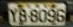}} \hspace{1.5mm}
	}
	
    \vspace{2.25mm}
	
	\resizebox{\linewidth}{!}{
	\subfloat[][\centering \phantom{\,} \resizebox{\adj}{!}{\textbf{\phantom{1}\cite{masood2017sighthound}:}} \texttt{\textcolor{red}{I}51735}\hspace{\textwidth} \phantom{\,}  \resizebox{\adj}{!}{\textbf{Ours:}} \texttt{\textcolor{red}{I}51735}]{
		\includegraphics[width=0.23\linewidth]{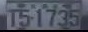}} \, %
	\subfloat[][\centering  \phantom{\,} \resizebox{\adj}{!}{\textbf{\phantom{1}\cite{masood2017sighthound}:}}~\texttt{9J3167}\hspace{\textwidth} \phantom{\,} \resizebox{\adj}{!}{\textbf{Ours:}}~\texttt{9J3167}]{
	\includegraphics[width=0.23\linewidth]{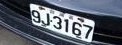}} \, %
	\subfloat[][\centering \phantom{\,} \resizebox{\adj}{!}{\textbf{\phantom{1}\cite{masood2017sighthound}:}} \texttt{\textcolor{red}{1}2N4202}\hspace{\textwidth} \phantom{\,}  \resizebox{\adj}{!}{\textbf{Ours:}} \texttt{\phantom{a}2N4202}]{
		\includegraphics[width=0.23\linewidth]{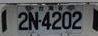}} \, %
    \subfloat[][\centering  \phantom{\,} \resizebox{\adj}{!}{\textbf{\phantom{1}\cite{masood2017sighthound}:}}~\texttt{\textcolor{red}{D}750J0}\hspace{\textwidth} \phantom{\,}  \resizebox{\adj}{!}{\textbf{Ours:}}~\texttt{0750J0}]{
		\includegraphics[width=0.23\linewidth]{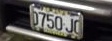}} \hspace{1.5mm}
	}
	
	\vspace{-0.5mm}

    \caption{\colored{Some qualitative results obtained on the \aolp~\cite{hsu2013application} dataset by Sighthound~\cite{masood2017sighthound} and the proposed system.}}
    \label{fig:detailed-aolp}
\end{figure}

\colored{
\noindent \textbf{\aolp~\cite{hsu2013application}:} this dataset has images collected \minor{in the Taiwan region} from front/rear views of vehicles and various locations, time, traffic, and weather conditions.
In our experiments, $683$ images were used for testing in each run.
As OpenALPR does not support \minor{\glspl*{lp} from the Taiwan region} (as pointed out in Section~\ref{sec:results:overall}), here we compare the results obtained by Sighthound~($87.1$\%) and the proposed system~($99.2$\%).
As shown in Fig.~\ref{fig:detailed-aolp}, different from what we expected, both systems dealt well with inclined \glspl*{lp} in this dataset.
While our system failed mostly in challenging cases, such as very similar characters (`E'~and~`F', `B'~and~`$8$', etc.), Sighthound also failed in simpler cases where our system had no difficulty in correctly recognizing all \gls*{lp}~characters.\\
}

\begin{figure}[!htb]
    \centering
    \captionsetup[subfigure]{labelformat=empty,font={scriptsize}}
    
    \resizebox{\linewidth}{!}{
	\subfloat[][\centering  \phantom{\,} \resizebox{\adj}{!}{\textbf{\phantom{1}\cite{masood2017sighthound}:}}~\texttt{BSE5579}\hspace{\textwidth} \phantom{\,} \resizebox{\adj}{!}{\textbf{\phantom{1}\cite{openalprapi}:}}~\texttt{BSE5579}\hspace{\textwidth} \phantom{\,}
    \resizebox{\adj}{!}{\textbf{\phantom{10}\cite{silva2018license}:}}~\texttt{BSE5579}\hspace{\textwidth} \phantom{\,} \resizebox{\adj}{!}{\textbf{Ours:}}~\texttt{BSE5579}]{
		\includegraphics[width=0.23\linewidth]{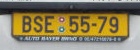}} \, %
    \subfloat[][\centering  \phantom{\,} \resizebox{\adj}{!}{\textbf{\phantom{1}\cite{masood2017sighthound}:}}~\texttt{BA2\textcolor{red}{2}8IM}\hspace{\textwidth} \phantom{\,} \resizebox{\adj}{!}{\textbf{\phantom{1}\cite{openalprapi}:}}~\texttt{BA268IM}\hspace{\textwidth} \phantom{\,}
    \resizebox{\adj}{!}{\textbf{\phantom{10}\cite{silva2018license}:}}~\texttt{\textcolor{red}{3}A268IM}\hspace{\textwidth} \phantom{\,} \resizebox{\adj}{!}{\textbf{Ours:}}~\texttt{BA268IM}]{
		\includegraphics[width=0.23\linewidth]{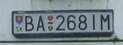}} \, %
	\subfloat[][\centering  \phantom{\,} \resizebox{\adj}{!}{\textbf{\phantom{1}\cite{masood2017sighthound}:}}~\texttt{VW4X4WP}\hspace{\textwidth} \phantom{\,} \resizebox{\adj}{!}{\textbf{\phantom{1}\cite{openalprapi}:}}~\texttt{VW4\textcolor{red}{7}4WP}\hspace{\textwidth} \phantom{\,}
    \resizebox{\adj}{!}{\textbf{\phantom{10}\cite{silva2018license}:}}~\texttt{VW4X4WP}\hspace{\textwidth} \phantom{\,} \resizebox{\adj}{!}{\textbf{Ours:}}~\texttt{VW4X4WP}]{
		\includegraphics[width=0.23\linewidth]{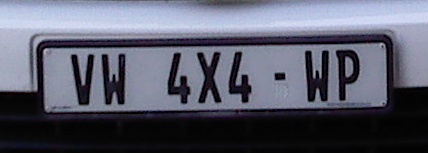}} \, %
    \subfloat[][\centering  \phantom{\,} \resizebox{\adj}{!}{\textbf{\phantom{1}\cite{masood2017sighthound}:}}~\texttt{GWAGEN}\hspace{\textwidth} \phantom{\,} \resizebox{\adj}{!}{\textbf{\phantom{1}\cite{openalprapi}:}}~\texttt{\textcolor{red}{-}WAGEN}\hspace{\textwidth} \phantom{\,}
    \resizebox{\adj}{!}{\textbf{\phantom{10}\cite{silva2018license}:}}~\texttt{G\textcolor{red}{VN}AGEN}\hspace{\textwidth} \phantom{\,} \resizebox{\adj}{!}{\textbf{Ours:}}~\texttt{G\textcolor{red}{VN}AGEN}]{
		\includegraphics[width=0.23\linewidth]{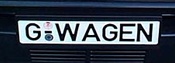}} \hspace{1.5mm}
	}
	
    \vspace{2.25mm}
	
	\resizebox{\linewidth}{!}{
    \subfloat[][\centering  \phantom{\,} \resizebox{\adj}{!}{\textbf{\phantom{1}\cite{masood2017sighthound}:}}~\texttt{WSQ3021}\hspace{\textwidth} \phantom{\,} \resizebox{\adj}{!}{\textbf{\phantom{1}\cite{openalprapi}:}}~\texttt{WS\textcolor{red}{0}3021}\hspace{\textwidth} \phantom{\,}
    \resizebox{\adj}{!}{\textbf{\phantom{10}\cite{silva2018license}:}}~\texttt{WSQ302\textcolor{red}{-}}\hspace{\textwidth} \phantom{\,} \resizebox{\adj}{!}{\textbf{Ours:}}~\texttt{WSQ3021}]{
		\includegraphics[width=0.23\linewidth]{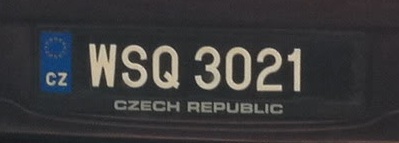}} \, %
	\subfloat[][\centering  \phantom{\,} \resizebox{\adj}{!}{\textbf{\phantom{1}\cite{masood2017sighthound}:}}~\texttt{RK60\textcolor{red}{0}AB}\hspace{\textwidth} \phantom{\,} \resizebox{\adj}{!}{\textbf{\phantom{1}\cite{openalprapi}:}}~\texttt{RK605AB}\hspace{\textwidth} \phantom{\,}
    \resizebox{\adj}{!}{\textbf{\phantom{10}\cite{silva2018license}:}}~\texttt{RK605AB}\hspace{\textwidth} \phantom{\,} \resizebox{\adj}{!}{\textbf{Ours:}}~\texttt{RK605AB}]{
		\includegraphics[width=0.23\linewidth]{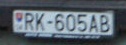}} \, %
	\subfloat[][\centering  \phantom{\,} \resizebox{\adj}{!}{\textbf{\phantom{1}\cite{masood2017sighthound}:}}~\texttt{1Z7\phantom{a}5233}\hspace{\textwidth} \phantom{\,} \resizebox{\adj}{!}{\textbf{\phantom{1}\cite{openalprapi}:}}~\texttt{1Z7\phantom{a}5233}\hspace{\textwidth} \phantom{\,}
    \resizebox{\adj}{!}{\textbf{\phantom{10}\cite{silva2018license}:}}~\texttt{1Z7\phantom{a}5233}\hspace{\textwidth} \phantom{\,} \resizebox{\adj}{!}{\textbf{Ours:}}~\texttt{1Z7\textcolor{red}{8}5233}]{
		\includegraphics[width=0.23\linewidth]{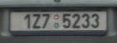}} \, %
    \subfloat[][\centering  \phantom{\,} \resizebox{\adj}{!}{\textbf{\phantom{1}\cite{masood2017sighthound}:}}~\texttt{RK161AG}\hspace{\textwidth} \phantom{\,} \resizebox{\adj}{!}{\textbf{\phantom{1}\cite{openalprapi}:}}~\texttt{\textcolor{red}{B}K161AG}\hspace{\textwidth} \phantom{\,}
    \resizebox{\adj}{!}{\textbf{\phantom{10}\cite{silva2018license}:}}~\texttt{RK161AG}\hspace{\textwidth} \phantom{\,} \resizebox{\adj}{!}{\textbf{Ours:}}~\texttt{RK161AG}]{
		\includegraphics[width=0.23\linewidth]{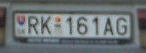}} \hspace{1.5mm}
	}
	
	\vspace{-0.5mm}

    \caption{\colored{Some qualitative results obtained on \openalpreu~\cite{openalpreu} by Sighthound~\cite{masood2017sighthound}, OpenALPR~\cite{openalprapi}, Silva \& Jung~\cite{silva2018license}, and the proposed system.}}
    \label{fig:detailed-openalpreu}
\end{figure}

\colored{
\noindent \textbf{\openalpreu~\cite{openalpreu}:} this dataset consists of $108$ testing images, generally with the vehicle well centered and occupying a large portion of the image.
Therefore, both our \gls*{alpr} system and the baselines performed well on this dataset.
Over five executions, the proposed system~($97.8$\%) failed in just $3$ different images, while the baselines failed in a few more.
Surprisingly, as can be seen in Fig.~\ref{fig:detailed-openalpreu}, the systems made distinct recognition errors and we were unable to find an explicit pattern among the incorrect predictions made by each of them.
In this sense, we believe that the errors in this dataset are mainly due to the great variability in the fonts of the characters in different \gls*{lp} layouts.
As an example, note in Fig.~\ref{fig:detailed-openalpreu} that the `W' character varies considerably depending on the \gls*{lp}~layout.
}

\begin{figure}[!htb]
    \centering
    \captionsetup[subfigure]{labelformat=empty,font={scriptsize}}
    
    \resizebox{\linewidth}{!}{
	\subfloat[][\centering  \phantom{\,} \resizebox{\adj}{!}{\textbf{\phantom{1}\cite{masood2017sighthound}:}}~\texttt{HIM2848}\hspace{\textwidth} \phantom{\,} \resizebox{\adj}{!}{\textbf{\phantom{1}\cite{openalprapi}:}}~\texttt{HIM2848}\hspace{\textwidth} \phantom{\,}
    \resizebox{\adj}{!}{\textbf{\phantom{10}\cite{silva2018license}:}}~\texttt{HIM2848}\hspace{\textwidth} \phantom{\,}
    \resizebox{\adj}{!}{\textbf{\phantom{1}\cite{laroca2018robust}:}}~\texttt{HIM2848}\hspace{\textwidth} \phantom{\,}
    \resizebox{\adj}{!}{\textbf{Ours:}}~\texttt{HIM2848}]{
		\includegraphics[width=0.23\linewidth]{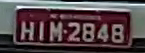}} \, %
    \subfloat[][\centering  \phantom{\,} \resizebox{\adj}{!}{\textbf{\phantom{1}\cite{masood2017sighthound}:}}~\texttt{OQ\textcolor{red}{D}541\textcolor{red}{D}}\hspace{\textwidth} \phantom{\,} \resizebox{\adj}{!}{\textbf{\phantom{1}\cite{openalprapi}:}}~\texttt{\textcolor{red}{D}QO5410}\hspace{\textwidth} \phantom{\,}
    \resizebox{\adj}{!}{\textbf{\phantom{10}\cite{silva2018license}:}}~\texttt{O\textcolor{red}{O}O5410}\hspace{\textwidth} \phantom{\,}
    \resizebox{\adj}{!}{\textbf{\phantom{1}\cite{laroca2018robust}:}}~\texttt{O\textcolor{red}{O}O5410}\hspace{\textwidth} \phantom{\,}
    \resizebox{\adj}{!}{\textbf{Ours:}}~\texttt{O\textcolor{red}{O}O5410}]{
		\includegraphics[width=0.23\linewidth]{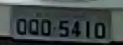}} \, %
	\subfloat[][\centering  \phantom{\,} \resizebox{\adj}{!}{\textbf{\phantom{1}\cite{masood2017sighthound}:}}~\texttt{H\textcolor{red}{O}R8361}\hspace{\textwidth} \phantom{\,} \resizebox{\adj}{!}{\textbf{\phantom{1}\cite{openalprapi}:}}~\texttt{HDR8361}\hspace{\textwidth} \phantom{\,}
    \resizebox{\adj}{!}{\textbf{\phantom{10}\cite{silva2018license}:}}~\texttt{HDR8361}\hspace{\textwidth} \phantom{\,}
    \resizebox{\adj}{!}{\textbf{\phantom{1}\cite{laroca2018robust}:}}~\texttt{HDR8361}\hspace{\textwidth} \phantom{\,}
    \resizebox{\adj}{!}{\textbf{Ours:}}~\texttt{H\textcolor{red}{O}R8361}]{
		\includegraphics[width=0.23\linewidth]{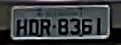}} \, %
    \subfloat[][\centering  \phantom{\,} \resizebox{\adj}{!}{\textbf{\phantom{1}\cite{masood2017sighthound}:}}~\texttt{HJN208\textcolor{red}{-}}\hspace{\textwidth} \phantom{\,} \resizebox{\adj}{!}{\textbf{\phantom{1}\cite{openalprapi}:}}~\texttt{\textcolor{red}{R}JN2081}\hspace{\textwidth} \phantom{\,}
    \resizebox{\adj}{!}{\textbf{\phantom{10}\cite{silva2018license}:}}~\texttt{H\textcolor{red}{L}N2081}\hspace{\textwidth} \phantom{\,}
    \resizebox{\adj}{!}{\textbf{\phantom{1}\cite{laroca2018robust}:}}~\texttt{HJN208\textcolor{red}{-}}\hspace{\textwidth} \phantom{\,}
    \resizebox{\adj}{!}{\textbf{Ours:}}~\texttt{HJN2081}]{
		\includegraphics[width=0.23\linewidth]{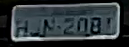}} \hspace{1.5mm}
	}
	
    \vspace{2.25mm}
	
	\resizebox{\linewidth}{!}{
	\subfloat[][\centering  \phantom{\,} \resizebox{\adj}{!}{\textbf{\phantom{1}\cite{masood2017sighthound}:}}~\texttt{\textcolor{red}{D}GQ6370}\hspace{\textwidth} \phantom{\,} \resizebox{\adj}{!}{\textbf{\phantom{1}\cite{openalprapi}:}}~\texttt{OG\textcolor{red}{O}6370}\hspace{\textwidth} \phantom{\,}
    \resizebox{\adj}{!}{\textbf{\phantom{10}\cite{silva2018license}:}}~\texttt{OG\textcolor{red}{O}6370}\hspace{\textwidth} \phantom{\,}
    \resizebox{\adj}{!}{\textbf{\phantom{1}\cite{laroca2018robust}:}}~\texttt{OG\textcolor{red}{O}6370}\hspace{\textwidth} \phantom{\,}
    \resizebox{\adj}{!}{\textbf{Ours:}}~\texttt{OG\textcolor{red}{O}6370}]{
		\includegraphics[width=0.23\linewidth]{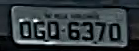}} \, %
	\subfloat[][\centering  \phantom{\,} \resizebox{\adj}{!}{\textbf{\phantom{1}\cite{masood2017sighthound}:}}~\texttt{\textcolor{red}{D}XH86\textcolor{red}{J}7}\hspace{\textwidth} \phantom{\,} \resizebox{\adj}{!}{\textbf{\phantom{1}\cite{openalprapi}:}}~\texttt{OXH8617}\hspace{\textwidth} \phantom{\,}
    \resizebox{\adj}{!}{\textbf{\phantom{10}\cite{silva2018license}:}}~\texttt{OXH8617}\hspace{\textwidth} \phantom{\,}
    \resizebox{\adj}{!}{\textbf{\phantom{1}\cite{laroca2018robust}:}}~\texttt{OXH86\textcolor{red}{-}7}\hspace{\textwidth} \phantom{\,}
    \resizebox{\adj}{!}{\textbf{Ours:}}~\texttt{OXH8617}]{
		\includegraphics[width=0.23\linewidth]{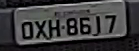}} \, %
	\subfloat[][\centering  \phantom{\,} \resizebox{\adj}{!}{\textbf{\phantom{1}\cite{masood2017sighthound}:}}~\texttt{\textcolor{red}{-}TV7556}\hspace{\textwidth} \phantom{\,} \resizebox{\adj}{!}{\textbf{\phantom{1}\cite{openalprapi}:}}~\texttt{\textcolor{red}{K}TV7556}\hspace{\textwidth} \phantom{\,}
    \resizebox{\adj}{!}{\textbf{\phantom{10}\cite{silva2018license}:}}~\texttt{\textcolor{red}{4}TV7556}\hspace{\textwidth} \phantom{\,}
    \resizebox{\adj}{!}{\textbf{\phantom{1}\cite{laroca2018robust}:}}~\texttt{A\textcolor{red}{I}V7556}\hspace{\textwidth} \phantom{\,}
    \resizebox{\adj}{!}{\textbf{Ours:}}~\texttt{ATV7556}]{
		\includegraphics[width=0.23\linewidth]{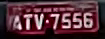}} \, %
    \subfloat[][\centering  \phantom{\,} \resizebox{\adj}{!}{\textbf{\phantom{1}\cite{masood2017sighthound}:}}~\texttt{GMF\textcolor{red}{-}862}\hspace{\textwidth} \phantom{\,} \resizebox{\adj}{!}{\textbf{\phantom{1}\cite{openalprapi}:}}~\texttt{GMF2862}\hspace{\textwidth} \phantom{\,}
    \resizebox{\adj}{!}{\textbf{\phantom{10}\cite{silva2018license}:}}~\texttt{G\textcolor{red}{N}F2862}\hspace{\textwidth} \phantom{\,}
    \resizebox{\adj}{!}{\textbf{\phantom{1}\cite{laroca2018robust}:}}~\texttt{G\textcolor{red}{N}F2862}\hspace{\textwidth} \phantom{\,}
    \resizebox{\adj}{!}{\textbf{Ours:}}~\texttt{GMF2862}]{
		\includegraphics[width=0.23\linewidth]{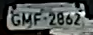}} \hspace{1.5mm}
	}
	
	\vspace{-0.5mm}

    \caption{\colored{Some qualitative results obtained on \ssig~\cite{goncalves2016benchmark} by Sighthound~\cite{masood2017sighthound}, OpenALPR~\cite{openalprapi}, Silva \& Jung~\cite{silva2018license}, the preliminary version of our approach~\cite{laroca2018robust}, and the proposed system.}}
    \label{fig:detailed-ssig}
\end{figure}

\colored{
\noindent \textbf{\ssig~\cite{goncalves2016benchmark}:} this dataset contains $800$ images for testing.
All images were taken with a static camera on the campus of a Brazilian university.
Here, the proposed system achieved a high recognition rate of~$98.2$\%, outperforming the best baseline by~$6.2$\%.
As shown in Fig.~\ref{fig:detailed-ssig}, as well as in other datasets, our system failed mostly in challenging cases where one character becomes very similar to another due to motion blur, the position of the camera, and other factors.
This was also the reason for most of the errors made by OpenALPR and the system designed by Silva \& Jung~\cite{silva2018license}.
However, these systems also struggled to correctly recognize degraded~\glspl*{lp} in which some characters are distorted or erased.
In addition to such errors, Sighthound predicted $6$ characters instead of $7$ on several occasions, probably because it does not take advantage of information regarding the \gls*{lp}~layout.
Lastly, the preliminary version of our approach~\cite{laroca2018robust}, where the \gls*{lp} characters are first segmented and then individually recognized, had difficulty segmenting the characters `I'~and~`$1$' in some cases, which resulted in recognition~errors.
}

\begin{figure}[!htb]
    \centering
    \captionsetup[subfigure]{labelformat=empty,font={scriptsize}}
    
    \resizebox{\linewidth}{!}{
	\subfloat[][\centering  \phantom{\,} \resizebox{\adj}{!}{\textbf{\phantom{1}\cite{masood2017sighthound}:}}~\texttt{ABN8528}\hspace{\textwidth} \phantom{\,} \resizebox{\adj}{!}{\textbf{\phantom{1}\cite{openalprapi}:}}~\texttt{ABN8528}\hspace{\textwidth} \phantom{\,}
    \resizebox{\adj}{!}{\textbf{\phantom{1}\cite{laroca2018robust}:}}~\texttt{ABN8528}\hspace{\textwidth} \phantom{\,} \resizebox{\adj}{!}{\textbf{Ours:}}~\texttt{ABN8528}]{
		\includegraphics[width=0.23\linewidth]{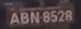}} \, %
    \subfloat[][\centering  \phantom{\,} \resizebox{\adj}{!}{\textbf{\phantom{1}\cite{masood2017sighthound}:}}~\texttt{AM\textcolor{red}{DD}663}\hspace{\textwidth} \phantom{\,} \resizebox{\adj}{!}{\textbf{\phantom{1}\cite{openalprapi}:}}~\texttt{AM\textcolor{red}{D}0663}\hspace{\textwidth} \phantom{\,}
    \resizebox{\adj}{!}{\textbf{\phantom{1}\cite{laroca2018robust}:}}~\texttt{AMO0663}\hspace{\textwidth} \phantom{\,} \resizebox{\adj}{!}{\textbf{Ours:}}~\texttt{AMO0663}]{
		\includegraphics[width=0.23\linewidth]{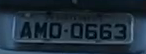}} \, %
	\subfloat[][\centering  \phantom{\,} \resizebox{\adj}{!}{\textbf{\phantom{1}\cite{masood2017sighthound}:}}~\texttt{ATT4025}\hspace{\textwidth} \phantom{\,} \resizebox{\adj}{!}{\textbf{\phantom{1}\cite{openalprapi}:}}~\texttt{ATT4025}\hspace{\textwidth} \phantom{\,}
    \resizebox{\adj}{!}{\textbf{\phantom{1}\cite{laroca2018robust}:}}~\texttt{AT\textcolor{red}{U}4025}\hspace{\textwidth} \phantom{\,} \resizebox{\adj}{!}{\textbf{Ours:}}~\texttt{ATT402\textcolor{red}{6}}]{
		\includegraphics[width=0.23\linewidth]{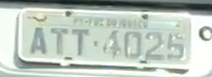}} \, %
    \subfloat[][\centering  \phantom{\,} \resizebox{\adj}{!}{\textbf{\phantom{1}\cite{masood2017sighthound}:}}~\texttt{AUG\textcolor{red}{-}936}\hspace{\textwidth} \phantom{\,} \resizebox{\adj}{!}{\textbf{\phantom{1}\cite{openalprapi}:}}~\texttt{A\textcolor{red}{D}G0936}\hspace{\textwidth} \phantom{\,}
    \resizebox{\adj}{!}{\textbf{\phantom{1}\cite{laroca2018robust}:}}~\texttt{AU\textcolor{red}{S}0936}\hspace{\textwidth} \phantom{\,} \resizebox{\adj}{!}{\textbf{Ours:}}~\texttt{AUG0936}]{
		\includegraphics[width=0.23\linewidth]{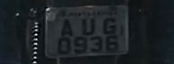}} \hspace{1.5mm}
	}
	
    \vspace{2.25mm}
	
	\resizebox{\linewidth}{!}{
    \subfloat[][\centering  \phantom{\,} \resizebox{\adj}{!}{\textbf{\phantom{1}\cite{masood2017sighthound}:}}~\texttt{BBO851\textcolor{red}{-}}\hspace{\textwidth} \phantom{\,} \resizebox{\adj}{!}{\textbf{\phantom{1}\cite{openalprapi}:}}~\texttt{BBO8514}\hspace{\textwidth} \phantom{\,}
    \resizebox{\adj}{!}{\textbf{\phantom{1}\cite{laroca2018robust}:}}~\texttt{BBO85\textcolor{red}{-}4}\hspace{\textwidth} \phantom{\,} \resizebox{\adj}{!}{\textbf{Ours:}}~\texttt{BBO8514}]{
		\includegraphics[width=0.23\linewidth]{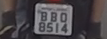}} \, %
	\subfloat[][\centering  \phantom{\,} \resizebox{\adj}{!}{\textbf{\phantom{1}\cite{masood2017sighthound}:}}~\texttt{IO\textcolor{red}{2}3616}\hspace{\textwidth} \phantom{\,} \resizebox{\adj}{!}{\textbf{\phantom{1}\cite{openalprapi}:}}~\texttt{IOZ3616}\hspace{\textwidth} \phantom{\,}
    \resizebox{\adj}{!}{\textbf{\phantom{1}\cite{laroca2018robust}:}}~\texttt{IOZ3616}\hspace{\textwidth} \phantom{\,} \resizebox{\adj}{!}{\textbf{Ours:}}~\texttt{IOZ3616}]{
		\includegraphics[width=0.23\linewidth]{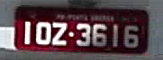}} \, %
	\subfloat[][\centering  \phantom{\,} \resizebox{\adj}{!}{\textbf{\phantom{1}\cite{masood2017sighthound}:}}~\texttt{A\textcolor{red}{O}W1379}\hspace{\textwidth} \phantom{\,} \resizebox{\adj}{!}{\textbf{\phantom{1}\cite{openalprapi}:}}~\texttt{\textcolor{red}{N}QW1379}\hspace{\textwidth} \phantom{\,}
    \resizebox{\adj}{!}{\textbf{\phantom{1}\cite{laroca2018robust}:}}~\texttt{\textcolor{red}{-}OW\textcolor{red}{7}379}\hspace{\textwidth} \phantom{\,} \resizebox{\adj}{!}{\textbf{Ours:}}~\texttt{A\textcolor{red}{O}W1379}]{
		\includegraphics[width=0.23\linewidth]{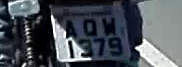}} \, %
	\subfloat[][\centering  \phantom{\,} \resizebox{\adj}{!}{\textbf{\phantom{1}\cite{masood2017sighthound}:}}~\texttt{AIQ\textcolor{red}{-Q}56}\hspace{\textwidth} \phantom{\,} \resizebox{\adj}{!}{\textbf{\phantom{1}\cite{openalprapi}:}}~\texttt{A\textcolor{red}{T}Q1056}\hspace{\textwidth} \phantom{\,}
    \resizebox{\adj}{!}{\textbf{\phantom{1}\cite{laroca2018robust}:}}~\texttt{A\textcolor{red}{UC}1056}\hspace{\textwidth} \phantom{\,} \resizebox{\adj}{!}{\textbf{Ours:}}~\texttt{A\textcolor{red}{T}Q1056}]{
		\includegraphics[width=0.23\linewidth]{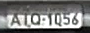}} \hspace{1.5mm}
	}
	
	\vspace{-0.5mm}

    \caption{\colored{Some qualitative results obtained on \dataset~\cite{laroca2018robust} by Sighthound~\cite{masood2017sighthound}, OpenALPR~\cite{openalprapi}, the preliminary version of our approach~\cite{laroca2018robust}, and the proposed system.}}
    \label{fig:detailed-ufpralpr}
\end{figure}

 \vspace{1.5mm}
\colored{
\noindent \textbf{\dataset~\cite{laroca2018robust}:} this challenging dataset includes $1{,}800$ testing images acquired from inside a vehicle driving through regular traffic in an urban environment, that is, both the vehicles and the camera (inside another vehicle) were moving and most \glspl*{lp} occupy a very small region of the image.
In this sense, the commercial systems did not return any prediction in some images from this dataset where the vehicles are far from the camera.
Regarding the recognition errors, they are very similar to those observed in the \ssig dataset.
Sighthound often confused similar letters and digits, while segmentation failures impaired the results obtained by the approach proposed in our previous work~\cite{laroca2018robust}.
According to Fig.~\ref{fig:detailed-ufpralpr}, the images were collected under different lighting conditions and the four \gls*{alpr} systems found it difficult to correctly recognize certain \glspl*{lp} with shadows or high exposure.
It should be noted that motorcycle \glspl*{lp} (those with two rows of characters) are challenging in nature, as the characters are smaller and closely spaced.
In this context, some authors have evaluated their methods, which do not work for motorcycles or for \glspl*{lp} with two rows of characters, exclusively in images containing cars, overlooking those with motorcycles~\cite{goncalves2018realtime,silva2020realtime}.
}

\vspace{1.5mm}
\colored{
\noindent \textbf{Final remarks:} while being able to process in real time, the proposed system is also capable of correctly recognizing \glspl*{lp} from several countries/regions in images taken under different conditions.
In general, our \gls*{alpr} system failed in challenging cases where one character becomes very similar to another due to factors such as shadows and occlusions (note that some of the baselines also failed in most of these cases).
We believe that vehicle information, such as make and model, can be explored in our system's pipeline in order to make it even more robust and prevent errors in such~cases.
}
\section{Conclusions}
\label{sec:conclusions}

\minor{In this work, as our main contribution, we presented an end-to-end, efficient and layout-independent \gls*{alpr} system that explores YOLO-based models at all stages.}
The proposed system contains a unified approach for \gls*{lp} detection and layout classification to improve the recognition results using post-processing rules.
\minor{This strategy proved essential for reaching outstanding results} since, depending on the \gls*{lp} layout, we avoided errors in characters that are often misclassified and also in the number of predicted characters to be~considered.

Our system achieved an average recognition rate of \accavg across eight public datasets used in the experiments, outperforming Sighthound and OpenALPR by \outsighthound and \outopenalpr, respectively.
More specifically, the proposed system outperformed both previous works and commercial systems in the \chinese, \openalpreu, \ssig and \dataset datasets, and yielded competitive results to those attained by the~baselines in the other datasets. 

We also carried out experiments to measure the execution time. 
Compared to previous works, our system achieved an impressive trade-off between accuracy and speed.
Specifically, even though the proposed approach achieves high recognition rates (i.e., above $95$\%) in all datasets except \dataset (where it outperformed the best baseline by $7.8$\%), it is able to process images in real time even when there are $4$ vehicles in the~scene. 
\minor{In this sense, we believe that our \gls*{alpr} system can run fast enough even in mid-end setups/GPUs.}

Another important contribution is that we manually labeled the position of the vehicles, \glspl*{lp} and characters, as well as their classes, in all datasets used in this work that have no annotations or that contain labels only for part of the \gls*{alpr} pipeline.
Note that the labeling process took a considerable amount of time since there are several bounding boxes to be labeled on each image (precisely, we manually labeled \boundingboxes bounding boxes on \images images).
These annotations are \textit{publicly available} to the research community, assisting the development and evaluation of new \gls*{alpr} approaches as well as the fair comparison among published~works.

\minor{
We remark that the proposed system can be exploited in several applications in the context of intelligent transportation systems.
For example, it can clearly help re-identify vehicles of the same model and color in non-overlapping cameras through \gls*{lp} recognition~\cite{oliveira2019vehicle} --~note that very similar vehicles can be easily distinguished if they have different \gls*{lp} layouts.
Considering the impressive results achieved for  \gls*{lp} detection, it can also be explored for the protection of privacy in images obtained in urban environments by commercial systems such as \textit{Mapillary} and \textit{Google Street~View}~\cite{uittenbogaard2019privacy}.
}

As future work, we intend to design new \gls*{cnn} architectures to further optimize (in terms of speed) vehicle detection.
We also plan to correct the alignment of the detected \glspl*{lp} and also rectify them in order to achieve even better results in the \gls*{lp} recognition stage. 
\minor{Finally, we want to investigate the impact of various factors (e.g., concurrent processes, hardware characteristics, frameworks/libraries used, among others) on real-time processing thoroughly.
Such an investigation is of paramount importance for real-world applications, but it has not been done in the \gls*{alpr}~literature.
}
\section*{Acknowledgments}
This work was supported by the National Council for Scientific and Technological Development~(CNPq) (grant numbers~311053/2016-5, 428333/2016-8, 313423/2017-2 and 438629/2018-3); the Foundation for Research of the State of Minas Gerais~(FAPEMIG) (grant numbers~APQ-00567-14 and PPM-00540-17); and the Coordination for the Improvement of Higher Education Personnel~(CAPES) (Social Demand Program and DeepEyes Project).
The Titan~Xp used for this research was donated by the NVIDIA~Corporation.
%

\scriptsize
\balance
\setlength{\bibsep}{3pt}
\bibliographystyle{IEEEbib}
\bibliography{bibtex}

\end{document}